\documentclass[10pt,twocolumn,letterpaper]{article}

\usepackage[pagenumbers]{cvpr} 

\usepackage[dvipsnames]{xcolor}
\newcommand{\red}[1]{{\color{red}#1}}

\definecolor{cvprblue}{rgb}{0.21,0.49,0.74}
\usepackage[pagebackref,breaklinks,colorlinks,citecolor=cvprblue]{hyperref}

\usepackage[capitalize]{cleveref}
\crefname{section}{Sec.}{Secs.}
\Crefname{section}{Section}{Sections}
\Crefname{table}{Table}{Tables}
\crefname{table}{Tab.}{Tabs.}

\def\confName{CVPR}
\def\confYear{2024}

\usepackage{color}
\usepackage{xcolor}
\definecolor{deepGreen}{RGB}{0,153,0}
\definecolor{orange}{RGB}{255,125,0}
\def\red#1{\textcolor[rgb]{1,0,0}{#1}}
\def\blue#1{\textcolor[rgb]{0,0,1}{#1}}
\def\limegreen#1{\textcolor[rgb]{0,1,0}{#1}}
\def\green#1{\textcolor[rgb]{0.23,0.7,0.25}{#1}}
\def\truegreen#1{\textcolor[rgb]{0.6,0.81,0.47}{#1}}
\def\skyblue#1{\textcolor[rgb]{0.49,0.65,0.87}{#1}}

\newcommand{\keypoint}[1]{\vspace{0.1cm}\noindent\textbf{#1}\;}
\newcommand{\cut}[1]{}
\usepackage{multirow}
\usepackage{amsmath}
\usepackage{amssymb}
\usepackage{mathtools}
\usepackage{arydshln}
\usepackage{soul}
\usepackage{nicefrac}
\usepackage{booktabs}
\usepackage{stmaryrd}
\usepackage{caption}
\usepackage{graphicx}
\usepackage{booktabs}
\usepackage{colortbl}
\definecolor{Gray}{gray}{0.9}
\definecolor{pink}{RGB}{255, 234, 232}
\definecolor{good}{RGB}{214, 232, 212}
\definecolor{reasonable}{RGB}{188, 200, 211}
\definecolor{abstract}{RGB}{242, 214, 213}
\definecolor{columbiablue}{rgb}{0.61, 0.87, 1.0}
\def\graytext#1{\textcolor[RGB]{85,85,85}{#1}}
\usepackage{algorithm}
\usepackage{algpseudocode}
\definecolor{commentcolor}{RGB}{110,154,155}   
\newcommand{\PyComment}[1]{\ttfamily\textcolor{commentcolor}{\# #1}}  
\newcommand{\PyCode}[1]{\ttfamily\textcolor{black}{#1}} 

\usepackage[accsupp]{axessibility}

\newcommand{\MYhref}[3][blue]{\href{#2}{\color{#1}{#3}}}
\makeatletter
\apptocmd\@maketitle{{\myfigure{}\par}}{}{}
\makeatother
\usepackage{capt-of,etoolbox}

\title{\vspace{-0.7cm}How to Handle \emph{Sketch-Abstraction} in Sketch-Based Image Retrieval?\vspace{-0.5cm}}

\author{\MYhref[cvprblue]{https://subhadeepkoley.github.io}{Subhadeep Koley}\textsuperscript{1,2} \hspace{.2cm} \MYhref[cvprblue]{https://ayankumarbhunia.github.io}{Ayan Kumar Bhunia}\textsuperscript{1} \hspace{.2cm} \MYhref[cvprblue]{https://aneeshan95.github.io}{Aneeshan Sain}\textsuperscript{1} \hspace{.2cm}  \MYhref[cvprblue]{https://www.pinakinathc.me}{Pinaki Nath Chowdhury}\textsuperscript{1} \\ \MYhref[cvprblue]{https://www.surrey.ac.uk/people/tao-xiang}{Tao Xiang}\textsuperscript{1,2} \hspace{.2cm} \MYhref[cvprblue]{https://www.surrey.ac.uk/people/yi-zhe-song}{Yi-Zhe Song}\textsuperscript{1,2} \\
\textsuperscript{1}SketchX, CVSSP, University of Surrey, United Kingdom.  \\
\textsuperscript{2}iFlyTek-Surrey Joint Research Centre on Artificial Intelligence.\\
{\tt\small \{s.koley, a.bhunia, a.sain, p.chowdhury, t.xiang, y.song\}@surrey.ac.uk}\\
\small \url{https://subhadeepkoley.github.io/AbstractAway}
}
\vspace{-0.6cm}

\newcommand\myfigure{
\centering
\vspace{-0.9cm}
\captionsetup{type=figure} 
    \includegraphics[width=\textwidth]{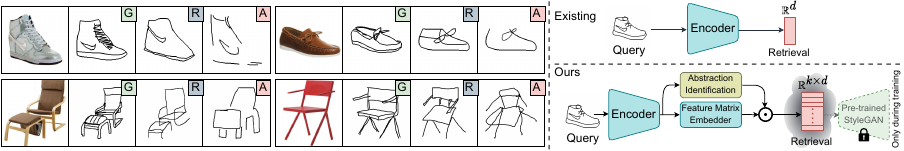}
    \vspace{-0.7cm}
\captionof{figure}{\textit{Left:} Freehand sketches exhibit varied levels of abstraction -- \colorbox{good}{\textbf{G}}: good, \colorbox{reasonable}{\textbf{R}}: reasonable, \colorbox{abstract}{\textbf{A}}: abstract. \textit{Right:} Unlike existing feature vector embedding, we learn a \textit{feature matrix} representation in the joint embedding space, \emph{regularised by a pre-trained StyleGAN's} disentangled latent space, and an \emph{abstraction-aware retrieval loss}. The \textit{abstraction identification} head dynamically decides the row-dimension of the matrix embedding based on the query sketch abstraction/completeness.
}

\label{fig:teaser}
\vspace{+0.2cm}
}

\begin{document}
\maketitle

\begin{abstract}
\vspace{-0.3cm}
   In this paper, we propose a novel abstraction-aware sketch-based image retrieval framework capable of handling sketch abstraction at varied levels. Prior works had mainly focused on tackling sub-factors such as drawing style and order, we instead attempt to model abstraction as a whole, and propose feature-level and retrieval granularity-level designs so that the system builds into its DNA the necessary means to interpret abstraction. On learning abstraction-aware features, we for the first-time harness the rich semantic embedding of pre-trained StyleGAN model, together with a novel abstraction-level mapper that deciphers the level of abstraction and dynamically selects appropriate dimensions in the feature matrix correspondingly, to construct a feature matrix embedding that can be freely traversed to accommodate different levels of abstraction. For granularity-level abstraction understanding, we dictate that the retrieval model should not treat all abstraction-levels equally and introduce a differentiable surrogate \texttt{Acc.@q} loss to inject that understanding into the system. Different to the gold-standard triplet loss, our \texttt{Acc.@q} loss uniquely allows a sketch to narrow/broaden its focus in terms of how stringent the evaluation should be -- the more abstract a sketch, the less stringent (higher $q$). Extensive experiments depict our method to outperform existing state-of-the-arts in standard SBIR tasks along with challenging scenarios like early retrieval, forensic sketch-photo matching, and style-invariant retrieval.
\end{abstract}

\vspace{-0.6cm}
\section{Introduction}
\vspace{-0.2cm}

Unquestionably the biggest difference between sketches and photos lies with that of abstraction -- sketches are abstract depictions, photos being lifelike~\cite{wei2021fine, koley2023picture, koley2023its}. Abstraction in sketches is formed by many interlacing factors such as drawing skill, style, culture and subjective interpretation~\cite{muhammad2018learning, bhunia2023sketch2saliency, chowdhury2023what, koley2023you, koley2024text}. As a result, human sketches typically exhibit a large variation in terms of abstraction-levels -- from good (art-trained), to reasonable (you), and those highly abstract (me!), as shown in \cref{fig:teaser}(left).

The sketch community had been on a quest to tackle abstraction since inception~\cite{muhammad2018learning, muhammad2019goal, yang2021sketchaa, alaniz2022abstracting, sain2021stylemeup}, however mainly focusing on one sub-element at once (\eg, style~\cite{sain2021stylemeup}). This paper attempts for the first time to tackle sketch abstraction as a whole, for the problem of fine-grained sketch-based image retrieval (FG-SBIR)~\cite{bhunia2020sketch, bhunia2021more, bhunia2022adaptive, sain2022sketch3t, chowdhury2022partially, sain2023clip, chowdhury2023scenetrilogy}. The end result is an abstraction-aware retrieval model that flexibly adapts to different levels of sketch abstraction while maintaining performance -- it takes the abstraction \textit{away}.

We operate under two guiding principles to tackle abstraction -- on feature level, and on retrieval granularity -- all to ensure our system has in its DNA means to accommodate all abstract forms of human sketches. On the former, we desire a flexible feature embedding that dynamically attends to the varying levels of abstraction. That is, under a highly abstract input sketch, we would like to return coarser features to conduct sketch-photo matching, and vice versa for those less abstract (more detailed). On the latter, we dictate that our system builds into its understanding, that retrieval granularity (\ie, how fine-grained the retrieval is) is \textit{negatively} correlated with sketch abstraction-level. This is intuitive -- one can not reasonably expect a rough contour sketch of a shoe, to magically retrieve a specific Jordan trainer.

To engineer the said abstraction-aware feature embedding, we \textit{(i)} make clever use of a pre-trained StyleGAN~\cite{karras2019style} model, which is shown via a pilot study~(\cref{sec:pilot}) to already exhibit a nicely disentangled coarse-to-fine latent feature embedding~\cite{yang2021semantic}, to guide the construction of an abstraction-aware feature embedding for FG-SBIR such that it follows the same coarse-to-fine pattern, and \textit{(ii)} design a novel abstraction-level mapper that first identifies the abstraction-level of a sketch, then dynamically selects dimensions in the feature embedding that correspond to the identified abstraction-level. These two designs elegantly interact with each other via a $k\times d$ feature \textit{matrix} embedding, other than the conventional column \textit{vector}~\cite{yu2016sketch}. It follows that by traversing across the rows of this matrix, one could easily vary the ``fine-grainedness'' of the sketch feature -- \textit{more} rows to encode \textit{less} abstract sketches.

Embedding an understanding of retrieval granularity is largely analogous to that of the focus of a camera\footnote{{\url{https://tinyurl.com/3tp3b732}}}, but for a sketch -- the more abstract the input sketch is, the less focused it should be. Translating to the retrieval setup, this means the system should be more tolerant given a highly abstract sketch, in terms of how well the sketch can perform retrieval, \ie, the sketch becomes less ``focused''. For that, we conduct another pilot study which shows the de facto triplet loss~\cite{yu2016sketch}, the gold standard in the FG-SBIR community~\cite{song2017deep,bhunia2021more,sain2022sketch3t,chowdhury2022partially, sain2023exploiting}, is not fit for purpose. This is intuitive as well since triplet loss is solely concerned with bringing the best match to top-1, regardless of how good the sketch is. Instead, we propose a \textit{differentiable} surrogate \texttt{Acc.@q} loss mimicking actual test-time evaluation of FG-SBIR. It follows that $q$ is our desired ``focus'' parameter -- larger the $q$, lesser the ``focus'' (retrieval granularity).

In summary, \textit{(i)} we for the first time tackle the abstraction problem as a whole for the task of FG-SBIR, \textit{(ii)} we tailor abstraction understanding in our system's DNA by specifically designing feature-level and retrieval granularity-level innovations, \textit{(iii)} we utilise the information-rich StyleGAN~\cite{karras2019style} latent space to guide learning of abstraction-aware feature embedding, \textit{(iv)} we propose a novel abstraction-level selector to dynamically assess levels of abstraction, and select appropriate levels in our feature matrix embedding, \textit{(v)} we develop a differentiable \texttt{Acc.@q} surrogate loss that allows the retrieval metric to accommodate different levels of abstraction. Comprehensive experiments show our method to surpass prior arts in a variety of FG-SBIR tasks such as early retrieval,  forensic sketch-photo matching, and style-invariant retrieval.

\vspace{-0.2cm}
\section{Related Works}
\vspace{-0.3cm}
\keypoint{Sketch-Based Image Retrieval.} Unlike category-level SBIR~\cite{collomosse2019livesketch, liu2017deep, dey2019doodle}, \textit{fine-grained}-SBIR (FG-SBIR) attempts at instance-specific sketch-photo correspondence \cite{sain2021stylemeup, bhunia2021more, bhunia2022sketching, yu2016sketch, xu2022domain, guo2017sketch, thakur2023active}. The deep Siamese triplet network~\cite{yu2016sketch} for FG-SBIR was further accelerated by attention-based higher-order loss~\cite{song2017deep}, text tags~\cite{song2017fine}, semi-supervised learning~\cite{bhunia2021more}, or self-supervised pre-training strategies~\cite{pang2020solving}.  Although subsequent works have tackled partial~\cite{bhunia2020sketch} and noisy~\cite{bhunia2022sketching} sketches, hierarchical abstraction~\cite{sain2020cross}, sketch-style diversity~\cite{sain2021stylemeup, bhunia2022adaptive}, the formal addressal of sketch abstraction in FG-SBIR remains unexplored till date.

\keypoint{Handling Sketch Abstraction.} \textit{Freehand sketch abstraction}~\cite{muhammad2018learning}, refers to the level of detail at which a sketch is represented~\cite{berger2013style} based on user's expertise, preference, time-limit, or the task in hand~\cite{muhammad2018learning}. Recent approaches for handling sketch abstraction can be broadly categorised as --
\textit{(a) Subset-selection}~\cite{muhammad2018learning, muhammad2019goal}, \textit{(b) Parametric sketch representation}~\cite{das2020beziersketch, das2021cloud2curve, das2021sketchode} \textit{(c) Primitive-based}~\cite{alaniz2022abstracting}, and \textit{(d) Abstraction-guided deep encoder}~\cite{yang2021sketchaa}. Under selection-based methods, Muhammad \etal~\cite{muhammad2018learning, muhammad2019goal} used the hard assumption that any freehand sketch can be crudely approximated from a subset containing the \textit{most salient} contours of the corresponding edgemap. Compared to $2D$ coordinates~\cite{bhunia2021vectorization, bhunia2021more}, recent works focus on learning a better representation of strokes through \textit{parametric} curves via B\'ezier control points~\cite{das2020beziersketch, vinker2022clipasso}, or differential equations~\cite{das2021sketchode}. These methods modulate the degree of abstraction (smoothness of the curve) by either varying the cardinality of B\'ezier control point~\cite{das2020beziersketch} or the sinusoidal frequency parameters of a differential equation~\cite{das2021sketchode}. Contrarily, shape-based methods~\cite{alaniz2022abstracting} handle abstraction by representing a sketch with a predefined set of geometric \textit{primitives}. Finally, instead of a standard ImageNet pre-trained backbone, Yang \etal~\cite{yang2021sketchaa} designed a sketch-specific \textit{deep encoder} for granular learning of shape and appearance abstraction at varying levels. While the existing methods are driven by some heuristic~\cite{alaniz2022abstracting} or hard assumption~\cite{muhammad2018learning, muhammad2019goal}, we aim to model the abstraction for FG-SBIR utilising the smooth latent space~\cite{xu2021generative} of a pre-trained StyleGAN~\cite{karras2019style} and a parametric surrogate loss.

\keypoint{GAN for Vision Task.} Primarily aimed at image generation, GAN has now become the de facto choice for various downstream tasks~\cite{fox2021stylevideogan, yu2018generative, zhu2016generative, abdal2021labels4free}. Delving deeper, it was revealed that the layer-wise representations~\cite{yang2021semantic, xu2021generative} learned by StyleGAN~\cite{karras2019style} or BigGAN~\cite{brock2018large} disentangle object semantics at different abstraction hierarchy (\ie, fine, medium, and coarse). Built on this disentanglement, GAN-inversion has paved the way for several downstream tasks \cite{alaluf2021restyle, richardson2021encoding, alaluf2021only}. While GAN has various sketch-related applications in image synthesis~\cite{chen2018sketchygan, wang2021sketch}, image/video editing~\cite{zeng2022sketchedit, yang2020deep, DeepFaceVideoEditing2022}, sketch completion~\cite{liu2019sketchgan}, we exploit the disentanglement property of a pre-trained StyleGAN~\cite{karras2019style} to model sketch abstraction for FG-SBIR.

\keypoint{Loss Functions for FG-SBIR.}
FG-SBIR practically aims at ranking all photos of a gallery by their proximity to the query-sketch~\cite{yu2016sketch}, with its paired photo ideally topping the rank list. Although test-time evaluation metrics (\eg, Recall@q, Precision@q, Accuracy@q) are usual for FG-SBIR, optimising them directly via gradient descent is difficult in practice, as they invoke non-differentiability~\cite{brown2020smooth, patel2022recall}, owing to operations like sorting~\cite{patel2022recall}, counting, etc. Consequently, state-of-the-arts~\cite{yu2016sketch, sain2021stylemeup, sain2022sketch3t} mostly employ triplet-based contrastive proxy loss that minimises the distance between anchor-sketch and paired positive photo, compared to a random negative photo. While a plethora of deep metric losses~\cite{hadsell2006dimensionality, wu2017sampling, oh2016deep, wen2016discriminative} exist, triplet loss~\cite{schroff2015facenet} has become a standard baseline owing to its ease of training, and performance~\cite{yu2016sketch}. This was enhanced further with a higher-order loss~\cite{song2017deep}, quadruplet loss~\cite{song2017fine}, meta-learning its margin value~\cite{bhunia2022adaptive}, or moving to a reinforcement learning (RL) based pipeline~\cite{bhunia2020sketch, bhunia2022sketching} that optimises rank (non-differentiable) of the paired photo. RL being allegedly unstable~\cite{bhunia2020sketch}, we take to \textit{surrogate loss} literature~\cite{brown2020smooth} that aims for a \textit{differentiable approximation} of evaluation metrics~\cite{brown2020smooth}, to directly train a model. We thus leverage latent space of a pre-trained StyleGAN along with a differentiable approximation of retrieval-rank via a parametric surrogate loss, to develop an abstraction-aware FG-SBIR framework.

\vspace{-0.2cm}
\section{Backgrounds}
\vspace{-0.2cm}
\keypoint{Baseline FG-SBIR Model.} Given an input image $\mathcal{I}\in\mathbb{R}^{H \times W \times 3}$, we obtain a $d$-dimensional $l_2$ normalised feature $ f_i = \mathcal{F}_b(\mathcal{I}) \in \mathbb{R}^d$ using an ImageNet pre-trained VGG-16~\cite{simonyan2014very} backbone network $\mathcal{F}_b$, sharing weights between sketch and photo branches. It trains over a triplet loss~\cite{weinberger2009distance} that aims to minimise the distance $\delta(\cdot, \cdot)$ of a sketch-feature ($f_s$) from its paired positive photo-feature ($f_{p}$), while increasing that from a random negative ($f_{n}$) photo-feature. With margin $\mu >0$, the triplet loss is formulated as:

\vspace{-0.65cm}
\begin{equation}
    \mathcal{L}_{\text{triplet}}=\mathtt{max}\{0,\mu+\delta(f_s,f_{p})-\delta(f_s,f_{n})\}
\end{equation}
\vspace{-0.65cm}

\keypoint{StyleGAN.} In traditional GAN~\cite{radford2015unsupervised}, a noise vector $z\in\mathcal{Z}$ of size $\mathbb{R}^d$ sampled from a Gaussian distribution is fed to the input layer of the generator to obtain an RGB image. Contrarily, StyleGAN~\cite{karras2019style} first transforms a sampled $z\in\mathcal{Z}$ of size $\mathbb{R}^d$ into an intermediate latent code~\cite{abdal2019image2stylegan} $w^+\in\mathcal{W^+}$ of size $\mathbb{R}^{k\times d}$ using a non-linear latent mapper $f:\mathcal{Z}\rightarrow\mathcal{W^+}$ (an 8-layer MLP), where $k$ depends on resolution ($M \times M$) of output image as $k = 2\log_2(M)-2$~\cite{karras2019style}. The synthesis network $\mathcal{G}(\cdot)$ starts with a learned constant tensor of size $\mathbb{R}^{4\times 4\times d}$ and contains several progressive resolution blocks, with each block having the sequence \texttt{conv3$\times$3} $\shortrightarrow$ \texttt{AdaIN} $\shortrightarrow$ \texttt{conv3$\times$3} $\shortrightarrow$ \texttt{AdaIN}~\cite{karras2019style}. Each latent vector $w^+_i$, upon passing through a common affine transformation layer $A$, generates the style $y$ = $(y_s, y_b)$ = $A(w_i^+)$. Using adaptive instance normalisation (AdaIN), $y$ controls specifics of the output image at the $i^{th}$ level of synthesis network~\cite{karras2019style} by modulating the feature-map $x$ as $\texttt{AdaIN}(x,y)=y_{s}\frac{x-\mu(x)}{\sigma(x)}+y_{b}$. Despite being trained in an unsupervised manner, StyleGAN's latent space holds disentangled object representation at multiple abstraction-levels~\cite{yang2021semantic}. Pertaining to FG-SBIR, as no hard ground-truth supervision exists to model varying abstraction-levels, we use this disentanglement potential in context of FG-SBIR.

\vspace{-0.1cm}
\section{Pilot Study: Problems and Solutions}\label{sec:pilot}
\begin{figure}[!htbp]
    \centering
    \includegraphics[width=\linewidth]{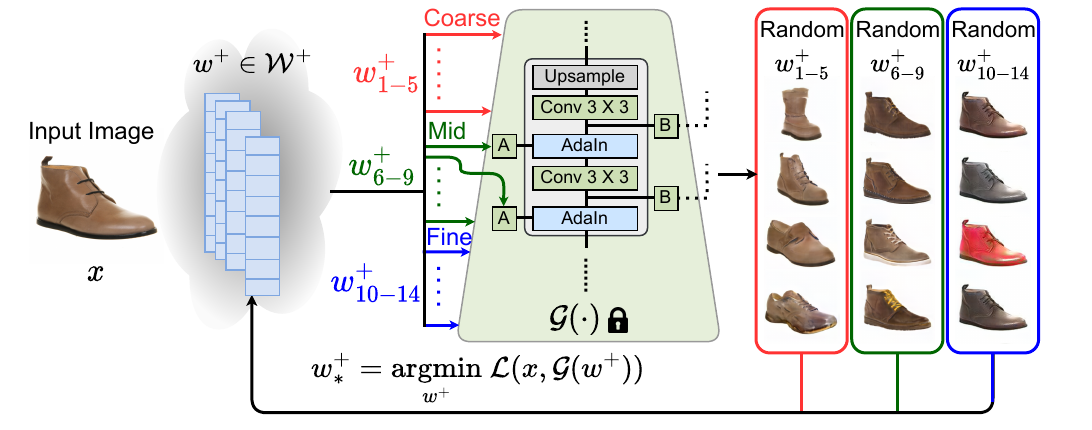}
    \vspace{-0.75cm}
    \caption{Pilot Study I: StyleGAN latent-disentanglement via optimising different groups of latent codes (\red{coarse}, \green{medium}, and \blue{fine}).}
    \label{fig:pilot_2}
    \vspace{-0.4cm}
\end{figure}

\vspace{-0.2cm}
\keypoint{How disentangled is the StyleGAN latent space?} We here aim to assess the extent of semantic feature disentanglement achievable via the StyleGAN latent code $w^+$ = $\{w^+_1, w^+_2, \cdots, w^+_k\}$ through a simple GAN inversion. In GAN inversion, given a real input image $x$ and an initial random latent code $w^+$, the aim is to find a new latent code $w^+_*$, that can most accurately reconstruct the input using a well-trained generator $\mathcal{G}(\cdot)$ via a reconstruction objective:  $w^+_*={\mathrm{argmin}_{w^+}}~\mathcal{L}(x,\mathcal{G}(w^+))$~\cite{abdal2019image2stylegan}. To figure out how different latent vectors in $w^+$ control certain specific attributes of the image, we segregate $w^+$ of size $\mathbb{R}^{14\times d}$  (for $256\times256$ real image) into three categories: $w^+_{coarse}$ = $\{w^+_1\rightarrow w^+_5\}$, $w^+_{mid}$ = $ \{w^+_6\rightarrow w^+_{9}\}$ and $w^+_{fine}$ = $\{w^+_{10}\rightarrow w^+_{14}\}$.  Given three latent groups $\{w^+_{coarse},w^+_{mid},w^+_{fine}\}$, we optimise \textit{any two} at once, while the third is sampled randomly from Gaussian distribution.
We consider all such \textit{exhaustive} latent group combinations ($^3C_2$) by repeating this experiment thrice with different seed values, which yields multiple reconstructed images as shown in \cref{fig:pilot_2}. Following existing literature~\cite{yang2021semantic}, we hypothesise that \emph{every group controls a specific set of attributes in the output image}, and not optimising a particular latent group (instead sampling from Gaussian distribution) will cause significant output variations corresponding to that set of attributes~\cite{yang2021semantic}. For instance, not optimising $w^+_{coarse}$ produces overall shape variation (\eg, shoe type), while doing so for $w^+_{mid}$, and $w^+_{fine}$ latent groups yield mid (\eg, lace, bootstrap) and fine (\eg, colour, appearance) level attribute variations, respectively.

This pilot study aims to motivate \textit{why} the disentangled latent representation of a pre-trained StyleGAN is a good choice for handling abstraction as a whole, whereas, later our proposed methodology (\cref{sec:method}) describes \textit{how} we use a pre-trained StyleGAN to learn an abstraction-aware feature matrix embedding in the context of FG-SBIR.

\vspace{-0.3cm}

\keypoint{Why model \textit{Abstraction} for FG-SBIR?} Highly abstract/partial early sketches may correspond randomly with multiple plausible photos from the gallery~\cite{bhunia2020sketch}, and as the drawing episode progresses towards completion, a sketch should start corresponding to a specific photo consistently~\cite{bhunia2020sketch}. We for the first time hypothesise this retrieval behaviour with a mathematical justification. Now, \textit{entropy} being a faithful indicator of randomness~\cite{wu2013local}, we use entropy of $\delta(\cdot,\cdot)$ in embedding space for a given query sketch $s^q_t$ over successive $(t\%=10\%, 20\%, \cdots, 100\%)$ levels of completion. Given a list of pre-computed photo features $\mathcal{P} = \{f_p^1, f_p^2, \cdots, f_p^N\}$, and a sketch query feature $f_s^t = \mathcal{F}(s_t^q)$ at $t\%$, we compute a pairwise distance list as $ \{\delta(f_s^t, f_p^1), \delta(f_s^t, f_p^2), \cdots, \delta(f_s^t, f_p^N)\}$. Similarly, the distance from paired positive photo feature $f_p^*$ is given as $\delta(f_s^t, f_p^*)$. We define the \textit{similarity} between sketch query and positive photo as a probability distribution over all the gallery photos at $t\%$ as:

\vspace{-0.8cm}
\begin{equation}
    p_t = {\mathrm{exp}(-\delta(f_s^t, f_p^*))} \bigg/{\textstyle\sum_{i=1}^{N} \mathrm{exp}(-\delta(f_s^t, f_p^i))}
    \vspace{-0.2cm}
\end{equation}

\begin{figure}[!htbp]
    \centering
    \includegraphics[width=\linewidth]{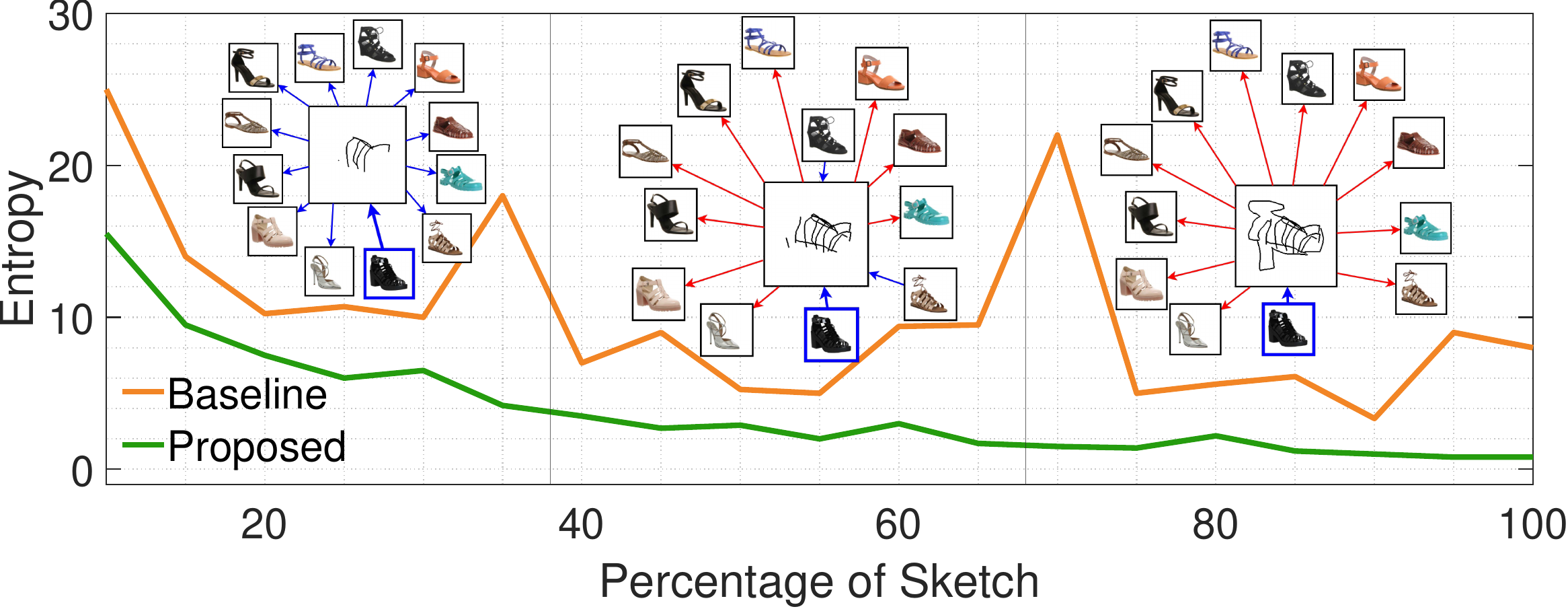}
    \vspace{-0.75cm}
    \caption{Pilot Study II: Compare retrieval consistency by comparing entropy of separation in the embedding space, evaluated over successive stages of sketch completion. Inset images show how our method directs the query to a single gallery image (\blue{blue}) while pushing others away as sketching progresses.}
    \label{fig:pilot_1}
    \vspace{-0.3cm}
\end{figure}

Following our hypothesis, as the sketch approaches completion ($t \rightarrow 100\%$), the distance between sketch query and positive photo should ideally decrease with respect to other non-matching gallery photos. This would lead to an increase in $p_t$, thus decreasing entropy $\mathcal{H}$ (= $-p_t\log p_t$). However, \cref{fig:pilot_1} shows that the baseline FG-SBIR model~\cite{yu2016sketch} deviates from our ideal hypothesis, depicting an arbitrary entropy behaviour, hampering performance. This calls for explicit abstraction-modelling for sketches. Accordingly, when this is addressed in our model, it depicts a \textit{consistent} entropy-decrease, thus following our hypothesis in guiding the sketch query towards a specific gallery instance as sketching reaches towards completion.

\vspace{-0.3cm}

\vspace{-0.3cm}
\section{Proposed Methodology}
\label{sec:method}
\vspace{-0.35cm}
\keypoint{Overview.} Our initial pilot study directs us towards two specific design elements: \textit{(i)} leveraging a pre-trained StyleGAN~\cite{karras2019style} during training, and \textit{(ii)} modelling the varied levels of sketch-abstraction via a novel loss function.
Firstly, compared to the embedding feature \textit{vector} of size $\mathbb{R}^d$ of baseline model~\cite{yu2016sketch}, we aim to use a feature \textit{matrix} embedding of size $\mathbb{R}^{k\times d}$ for sketch/photo representation in the joint-embedding space. In particular, we distil the disentangled knowledge~\cite{yang2021semantic} residing inside a pre-trained StyleGAN's~\cite{karras2020analyzing} latent space into our matrix embedding. Here specific rows of the matrix govern a few particular characteristics of the input, where moving down the rows {leads from} coarse to finer-level features~\cite{yang2021semantic}. Furthermore, our model automatically decides up to which row of the matrix embedding will be used for retrieval. Secondly, as triplet-based models~\cite{yu2016sketch} treat every sketch sample equally, ignoring their abstraction-level, we design a parametric surrogate loss that \textit{accounts} for \textit{individual} abstraction-level, thus leading towards a more practical training pipeline.

\vspace{-0.1cm}
\subsection{Model Architecture}
\vspace{-0.1cm}
The core of our model includes three major components: \textit{(i)} discovering which of the $\mathbb{R}^{k\times d}$ StyleGAN~\cite{karras2020analyzing} latent vectors are representative enough in context of FG-SBIR, \textit{(ii)} modifying the feature embedding network to generate \textit{matrix} instead of \textit{vector} embedding, \textit{(iii)} adaptively selecting the number of rows of the feature matrix embedding based on the input sketch abstraction for faithful retrieval.

\keypoint{Sketch-Specific Latent Groups.} Representing $256\times256$ images in StyleGAN~\cite{karras2019style} latent space would require latent codes of size $\mathbb{R}^{14\times d}$. As evident from our pilot study, initial latent vectors $\{w^+_{1}\rightarrow w^+_{9}\}$ control the major \textit{shape} and \textit{structure}, while the later ones $\{w^+_{10}\rightarrow w^+_{14}\}$ only govern the \textit{colour} and \textit{appearance} of the output. Likewise, \textit{sketch} being sparse binary line drawings, can only provide \textit{structural cues} in context of FG-SBIR~\cite{sain2021stylemeup}. Accordingly, during training, we exclusively use the structural latent vectors $\{w^+_{1}\rightarrow w^+_{9}\}$, resulting in a feature matrix embedding of size $\mathbb{R}^{9\times d}$. We further segregate the structural latent vectors into three sub-groups, each with a set of three latent vectors $\{w^+_{1\shortrightarrow3}, w^+_{4\shortrightarrow6}, w^+_{7\shortrightarrow9}\}$ to model a \emph{knob} for varying the level of sketch \underline{$a$}bstraction. For the rest of the paper, we name the groups as $a = \{a_{c}, a_{m}, a_{f}\}$ for the brevity of description.

\begin{figure}[!htbp]
    \centering
    \includegraphics[width=1\linewidth]{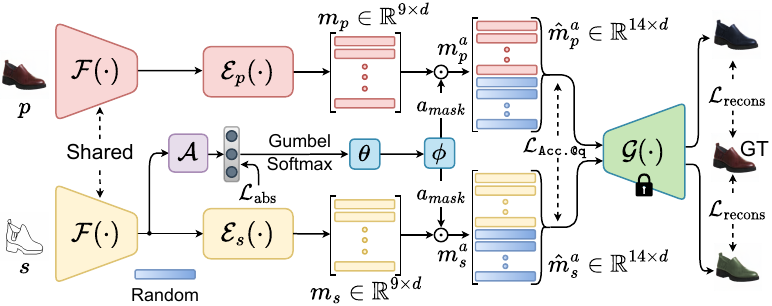}
    \vspace{-0.7cm}
    \caption{Our method learns a \textit{feature matrix} representation in the joint embedding space, regularised by a pre-trained StyleGAN, trained with a weighted summation of reconstruction, abstraction identification, and \texttt{Acc.@q} losses. $\theta$:$\texttt{flip}(\texttt{cumsum}(\texttt{flip}(\cdot)))$, $\phi$:$\texttt{repeat}~\text{and}~\texttt{flattening}~\text{operation}$ (more in text).}
    \label{fig:arch}
    \vspace{-0.6cm}
\end{figure}

\keypoint{Architecture.} A common backbone feature extractor $\mathcal{F}(\cdot)$ initialised from an ImageNet pre-trained VGG-16~\cite{simonyan2014very} is used to obtain a feature-map of size $\mathcal{F}(I) \in \mathbb{R}^{h \times w \times d}$ for both sketch ($s$) and photo ($p$) branches (\cref{fig:arch}). To tackle the large domain gap between pixel-perfect photos and sparse sketches~\cite{sain2020cross}, we apply individual sketch and photo specific \textit{feature matrix} embedding networks $\mathcal{E}_s$ and $\mathcal{E}_p$ on the output of $\mathcal{F}(I)$. Feature matrix embedding networks $\mathcal{E}_s, \mathcal{E}_p : \mathbb{R}^{h \times w \times d} \rightarrow \mathbb{R}^{9 \times d}$, each contains $9$-individual stride-two convolution blocks with LeakyReLU~\cite{xu2015empirical}, applied over $\mathcal{F}(I)$. Each of them predicts a $d$-dimensional feature, which upon concatenating across $9$-blocks gives the sketch feature matrix as $m_s = \mathcal{E}_s(\mathcal{F}(s))\in \mathbb{R}^{9 \times d}$ and photo feature matrix as $m_p = \mathcal{E}_p(\mathcal{F}(p))\in \mathbb{R}^{9 \times d}$, respectively.

\keypoint{Modelling Sketch Abstraction.} In context of FG-SBIR, we model sketch abstraction by dynamically deciding the number of row vectors from $m_s$ and $m_p$ to be used for distance calculation, where traversing from top to bottom acts as a \textit{tunable knob} controlling the degree of abstraction. Intuitively, for very coarse abstract sketches, we want to retrieve using only the $\{a_c\}$ group of latent vectors, while for sketches with medium and finer-grained details, we aim to retrieve using $\{a_c, a_m\}$ and $\{a_c, a_m, a_f\}$ groups respectively. Therefore, we add an abstraction identification head $\mathcal{A}: \mathbb{R}^d \rightarrow \mathbb{R}^{3}$ that takes a global average pooled sketch feature from the backbone network as input, and predicts the probability distribution over three latent groups as $\hat{a} =  p(a|s) = \mathcal{A}(\mathcal{F}(s))$ for a specific sketch input. After passing through a differentiable $\texttt{argmax}$ (realised with Gumbel-Softmax~\cite{jang2016categorical}), $\hat{a}$ gives a one-hot encoded vector depicting the abstraction-level. In order to generate the \emph{latent selection mask} differentiably, we apply cumulative summation as follows ${a}_{mask} = \texttt{flip}(\texttt{cumsum}(\texttt{flip}(\hat{a})))$, thus obtaining three possible mask states as $(\tilde{a}_c, \tilde{a}_m, \tilde{a}_f) \in \{ (1, 0, 0), (1, 1, 0), (1, 1, 1) \}$. In order to get the equivalent $9$-dimensional mask $\tilde{a}_{mask}$, we can simply use cascaded \texttt{repeat} and \texttt{flattening} operations\footnote{For example, $\{\red{1},\limegreen{1},\blue{0}\}\rightarrow\{\red{1},\red{1},\red{1},\limegreen{1},\limegreen{1},\limegreen{1},\blue{0},\blue{0},\blue{0}\}$} from any standard deep-learning library. Therefore given a sketch input, we can get the \textit{abstraction-aware} feature matrix embeddings as $m_s^a = m_s *\tilde{a}_{mask}$ and $m_p^a = m_p *\tilde{a}_{mask}$, where based on the predicted $\tilde{a}_{mask}$, the later rows of $m_p^a, m_s^a$ might get populated with zero values.

The question remains as to how we obtain the training signal to learn our abstraction identification head $\mathcal{A}$. Keeping the \textit{sequentially} abstract nature of sketches~\cite{bhunia2020sketch} in mind, we thus make a logical assumption of associating the coarse, medium and fine levels with vector sketches rendered at $30\%, 60\%$ and $100\%$ completion as one-hot encoded ground truth $a_{gt}$ for $\{a_c, a_m, a_f\}$ respectively.

\vspace{-0.05cm}
\subsection{Loss Objectives}
\vspace{-0.25cm}
\keypoint{Reconstruction Loss.}
To distil the knowledge from a pre-trained StyleGAN~\cite{karras2020analyzing}, we use sketch-to-photo and photo-to-photo reconstruction losses. We pad the $n$ non-zeros rows of ${m}_s^a$, and ${m}_p^a$ with $(14-n)$ $d$-dimensional vectors sampled from Gaussian distribution to obtain final matrix embeddings $\hat{m}_s^a$, and $\hat{m}_p^a$ of size $\mathbb{R}^{14\times d}$, as the pre-trained StyleGAN always expects a $\mathbb{R}^{14\times d}$ latent code to generate an output. Upon passing $\hat{m}_s^a$, and $\hat{m}_p^a$ through the pre-trained StyleGAN, we compute the reconstruction loss as:

\vspace{-0.3cm}
\begin{equation}
\mathcal{L}_{\text{recons}} =   \left \| p -  \mathcal{G}(\hat{m}_s^a)\right \|_2 +  \left \| p -  \mathcal{G}(\hat{m}_p^a)\right \|_2
\end{equation}
\vspace{-0.5cm}

Using this reconstruction objective, we aim to align the sketch/photo feature matrix embedding to that of the disentangled latent space of pre-trained StyleGAN~\cite{karras2020analyzing}, thus encouraging for a feature matrix where the abstraction-level decreases from \textit{coarse} to \textit{finer} across the rows.

\keypoint{Differentiable \texttt{Acc.@q} Retrieval Loss.}
Pre-trained StyleGAN~\cite{karras2019style}-based reconstruction loss alone, is not enough for discriminative learning in the feature matrix embedding space for cross-modal retrieval. As handling sketch abstraction is limited by triplet loss (\cref{sec:pilot}), we introduce a novel surrogate proxy loss \cite{patel2022recall, brown2020smooth} -- \texttt{Acc.@q}. Mimicking a standard evaluation-metric \cite{yu2016sketch}, \texttt{Acc.@q} measures the percentage of query samples having true match gallery photos appearing in the top-$q$ list. Varying $q$ as $10$, $5$, and $1$ helps to model coarse $(30\%)$, mid $(60\%)$ and fine $(100\%)$ levels of abstraction $\{a_c, a_m, a_f\}$ respectively. For instance, the initial \textit{coarse} sketch may correspond to multiple gallery photos where having the true match in the top-$10$ ($q$=$10$) list might suffice, whereas for a complete (\textit{fine}) sketch, we need to enforce retrieval at top-$1$ ($q$=$1$) position.

As computing rank across an entire dataset is impractical, we opt for a more reasonable choice of calculating \texttt{Acc.@q} across a batch. Given paired sketch-photo feature matrix embeddings across a batch with batch-size $B$ as $\{({m}_{s,i}^a, {m}_{p, i}^a)\}_{i=1}^{B}$, we first compute the pairwise-distances of a given sketch query ${m}_{s,i}^a$ with all the photos in a batch as $\mathcal{D}$ = $\{\delta_{s, i}^{p, 1}, \cdots,  \delta_{s, i}^{p, B}\}$, and true target distance with the paired photo as $\delta_{s, i}^{p, *}$. The relative distance vector now becomes $\mathcal{D}_{rel} = \delta_{s, i}^{p, *} - \mathcal{D}$. To calculate the rank of paired photo for $i^{th}$ sketch query, we need to find the number of positive values in $\mathcal{D}_{rel}^i$, which requires a \textit{non-differentiable} step function~\cite{patel2022recall, brown2020smooth} that we approximate via a temperature $(\tau)$ controlled Sigmoid function $S_{\tau}(x,\tau) = \frac{1}{1+\exp(-x/\tau)}$. Once we find $rank_i = \sum S_{\tau_2}(\mathcal{D}_{rel}^i)$, Acc.@q for that sketch query (either $+1$ or $0$) can easily be calculated  using $S_{\tau_1}(q - rank_i)$. Thus the overall \texttt{Acc.@q} loss becomes:

\vspace{-0.5cm}
\begin{equation}
    \mathcal{L}_\texttt{Acc.@q} = - \frac{\sum_{i=1}^{B} S_{\tau_1}(q -  \sum_{i=1}^{B} S_{\tau_2}(\mathcal{D}_{rel}^i))}{B}
    \vspace{-0.2cm}
\end{equation}

\noindent{A lower (\textit{higher}) value of $\tau$ leads to a better (\textit{worse}) approximation of the step function but results in a sparser (\textit{denser}) gradient~\cite{brown2020smooth}. Empirically, we find $\tau_1=1$ and $\tau_2=0.01$ to be optimum. A PyTorch-like pseudocode has been provided in \cref{alg:pseduCode} (for brevity, only $\mathbb{R}^d$ case is shown).}

\begin{figure}[!htbp]
\begin{algorithm}[H]
\scriptsize
\caption{PyTorch-like code for differentiable \texttt{Acc.@q}}\label{alg:pseduCode}
\begin{algorithmic}
\item \PyCode{\green{\textbf{import}} torch}
\item \PyCode{\green{\textbf{import}} torch.nn.functional \green{\textbf{as}} F}
\item \PyCode{\green{\textbf{import}} F.pairwise\_distance \green{\textbf{as}} PD}
\vspace{0.2cm}
\item \PyCode{\green{\textbf{def}} accuracy@q\_loss(anchor, pos, q):}
\item ~~~~\PyComment{anchor:B$\times$d; pos:B$\times$d; q:1$\times$1(Acc.@q)}
\item ~~~~\PyComment{B$\leftarrow$batch size; d$\leftarrow$feature dimension}
\item ~~~~\PyCode{acc\_q = torch.tensor(0)} \PyComment{Accumulate accuracy}
\item ~~~~\PyCode{dist = PD(anchor, pos)} \PyComment{B$\times$B}
\vspace{0.2cm}
\item ~~~~\PyCode{\green{\textbf{for}} i \green{\textbf{in}} range(len(anchor)):}
\item ~~~~~~~~\PyComment{Target distance for ${i^{th}}$ anchor}
\item ~~~~~~~~\PyCode{tgt\_dist = PD(anchor[i], pos[i])}\PyComment{1$\times$1}
\item ~~~~~~~~\PyComment{Relative distance to target}
\item ~~~~~~~~\PyCode{rel\_dist = tgt\_dist \textbf{-} PD(anchor[i], pos)}\PyComment{B$\times$1}
\vspace{0.2cm}
\item ~~~~~~~~\PyCode{signs = sigmoid(rel\_dist, 0.01)}
\item ~~~~~~~~\PyCode{rank\_i = torch.sum(signs)}\PyComment{Get rank:1$\times$1}
\item ~~~~~~~~\PyCode{acc\_q += sigmoid(q \textbf{-} rank\_i, 1)}
\vspace{0.2cm}
\item ~~~~\PyCode{batch\_acc = acc\_q / len(anchor)}
\item ~~~~\PyCode{loss = \textbf{-}1 * batch\_acc}
\item ~~~~\PyCode{\green{\textbf{return}} loss}
\vspace{0.2cm}
\item \PyComment{Approximate step function with Sigmoid}
\item \PyCode{\green{\textbf{def}} sigmoid(tensor, tau):}
\item ~~~~\PyCode{tensor = tensor / tau}
\item ~~~~\PyCode{y = 1.0 / (1.0 + torch.exp(\textbf{-}tensor))}
\item ~~~~\PyCode{\green{\textbf{return}} y}
\end{algorithmic}
\end{algorithm}
\vspace{-1cm}
\end{figure}

\keypoint{Abstraction Identification Loss.} Abstraction identification is a three-class classification problem that helps to \textit{dynamically decide} the degree of abstraction of the input sketch, thus dictating the cardinality of effective rows in the feature matrix embedding. Using the vector sketch rendered at $30\%$, $60\%$ or $100\%$ completion-levels as ground truth $a_{gt}$ for coarse, medium and fine categories respectively, our abstraction identification loss becomes:

\vspace{-0.4cm}
\begin{equation}
\label{eq:abs_identification}
    \mathcal{L}_{\text{abs}} = -\frac{1}{3}\sum_{i=1}^{3} a_{gt}^{i} \log(\hat{a}^i)
    \vspace{-0.2cm}
\end{equation}

In summary, our overall training objective can be defined as: $\mathcal{L}_{\text{total}}=\lambda_{1}\mathcal{L}_{\text{recons}} + \lambda_{2}\mathcal{L}_\texttt{Acc.@q} + \lambda_{3}\mathcal{L}_{\text{abs}}$. During inference, we \textit{discard} the StyleGAN~\cite{karras2020analyzing} generator module. The shared feature extractor ($\mathcal{F}$) followed by the feature matrix encoders ($\mathcal{E}_s$, $\mathcal{E}_p$) and the abstraction identification head ($\mathcal{A}$) is used to generate abstraction-aware feature matrix embeddings ${m}_s^a$ and ${m}_p^a$. We calculate the distance between ${m}_s^a$ and ${m}_p^a$ to perform the retrieval during inference.

\vspace{-0.06cm}
\section{Experiments}
\vspace{-0.25cm}
\keypoint{Dataset.} Following standard literature~\cite{sain2020cross,bhunia2020sketch}, we use QMUL-ShoeV2~\cite{bhunia2020sketch, song2018learning} and QMUL-ChairV2~\cite{bhunia2020sketch, song2018learning} datasets, to evaluate FG-SBIR performance. They contain $6730$ ($2000$) and $1800$ ($400$) sketches (photos) respectively, with multiple sketches per photo with fine-grained association. While $6051$ ($1275$) sketches and $1800$ ($300$) photos from ShoeV2 (ChairV2) are used for training respectively, the rest are used for testing. We use photos from the UT Zappos50K~\cite{yu2014fine} dataset for pre-training StyleGAN generator on the \textit{Shoe} class. Due to the lack of large-scale chair image datasets, we use over $10,000$ chair photos collected from shopping websites like IKEA, ARGOS, and Amazon to pre-train the StyleGAN generator on the \textit{Chair} class.

\keypoint{Implementation Details.} We use the Adam~\cite{kingma2014adam} optimiser to pre-train the StyleGAN generator for $8M$ iterations at a learning rate of $10^{-3}$ and a batch size of $8$. Empirically, we disable path-length regularisation and reduce $R1$ regularisation's weight to $2$ for stabler training with diverse output. We employ an ImageNet pre-trained VGG-16~\cite{simonyan2014very} as the feature backbone and a StyleGAN2~\cite{karras2020analyzing} model, both with an embedding dimension $d=512$. Our FG-SBIR model was trained for $3K$ epochs using the Adam~\cite{kingma2014adam} optimiser with a constant learning rate of $10^{-3}$ and batch size of $128$. Values of $\lambda_{1,2,3}$ are set to $0.5$, $1$, and $0.5$, empirically.

\keypoint{Evaluation.} Aligning with recent literature~\cite{bhunia2021more, chowdhury2022partially}, we evaluate our work on the standard FG-SBIR evaluation metric -- Acc.@q, which measures the percentage of sketches having a true-paired photo in the top-q retrieved list.

\keypoint{Human Study.} In this study, upon drawing an abstract sketch, human participants had to rate the top-ranked photo retrieved by every competing framework on a scale of $1$ to $5$ (bad$\shortrightarrow$excellent)~\cite{huynh2010study}, based on their \textit{opinion} of how closely the retrieved photo matched the \textit{imagination} of their previously drawn sketch. Each of the $72$ participants was asked to draw and rate $50$ such complete sketches, resulting in a total of $3600$ responses per method. For each method, we compute the final MOS value by taking the mean ($\mu$) and variance ($\sigma$) of all its MOS responses.

\keypoint{Competitors.} We evaluate the proposed method from three perspectives -- \textit{(i) Existing state-of-the-arts (SoTA):} \textbf{Triplet-SN}~\cite{yu2016sketch} uses Sketch-a-Net~\cite{yu2017sketch} backbone with standard triplet loss. \textbf{HOLEF-SN}~\cite{song2017deep} further extends \cite{yu2016sketch} with spatial attention-based higher-order loss. {\textit{(ii) \underline{B}aselines:} Owing to the lack of generative model-based FG-SBIR frameworks, we design three baselines. \textbf{B-pSp} first leverages a StyleGAN~\cite{karras2020analyzing}-based pSp~\cite{richardson2021encoding} model (pre-trained on ShoeV2 and ChairV2) to translate an input sketch to its photo domain. Next, it finds the matching feature in its nearest neighbourhood in the entire photo gallery using an ImageNet pre-trained VGG-16~\cite{simonyan2014very} feature extractor. \textbf{B-pix2pix} and \textbf{B-CycleGAN} follows the same paradigm as B-pSp, except using pre-trained pix2pix~\cite{isola2017image} and CycleGAN~\cite{zhu2017unpaired} models respectively for sketch-photo translation.} These baselines essentially transform \textit{sketch-based} image retrieval into an \textit{image-based} image retrieval problem. \textit{(iii) Partial/Style-invariant/Early FG-SBIR methods:} Here we compare the proposed method with recent state-of-the-art FG-SBIR methods like \textbf{Partial-OT}~\cite{chowdhury2022partially}, \textbf{Cross-Hier}~\cite{sain2020cross}, \textbf{StyleMeUP}~\cite{sain2021stylemeup}, \textbf{On-the-fly}~\cite{bhunia2020sketch}, and \textbf{SketchPVT}~\cite{sain2023exploiting}.

\vspace{-0.1cm}
\subsection{Quantitative Analysis}
\vspace{-0.1cm}
{Comparative results on FG-SBIR is delineated in \cref{tab:fgsbir}. Although HOLEF-SN \cite{song2017deep} performs better than Triplet-SN \cite{yu2016sketch} due to the adaptation of the attention module, both fare poorly when compared to ours, owing to their straightforward adaptation of a weaker backbone \cite{yu2017sketch}. {Generative baselines like B-pix2pix or B-CycleGAN fails to reach the performance of Triplet-SN \cite{yu2016sketch} due to the inefficiency of standard image translation models like pix2pix \cite{isola2017image} or CycleGAN \cite{zhu2017unpaired} in sketch-to-photo synthesis task \cite{liu2020unsupervised}. On the other hand, our method outperforms B-pix2pix and B-CycleGAN with a significant Acc.@1 margin of $18.6$ ($27.9$)$\%$ and $17.5$ ($27.0$)$\%$ in ShoeV2 (ChairV2) dataset respectively. B-pSp depicts sub-optimal Acc.@1 of $47.8$ $(30.1)\%$ in ShoeV2 (ChairV2) due to its naive adaptation of StyleGAN \cite{karras2020analyzing}.} Later methods attempt to address multiple different traits of freehand sketch like partiality \cite{chowdhury2022partially}, hierarchical-abstraction \cite{sain2020cross}, style-variation \cite{sain2021stylemeup}, early-retrieval \cite{bhunia2020sketch} for better accuracy. Surprisingly, our abstraction-aware model shows improved performance without the complicated optimal-transport of Partial-OT \cite{chowdhury2022partially}, costly hierarchical co-attention of CrossHier \cite{sain2020cross}, unstable meta-learning of StyleMeUp \cite{sain2021stylemeup}, or time-consuming RL-finetuning of On-the-fly \cite{bhunia2020sketch}. SketchPVT \cite{sain2023exploiting}, despite being the strongest replacement of ours, scores less than ours in every experimental setup. Furthermore, as discussed in the pilot study (\cref{sec:pilot}), our method achieves an average entropy of $3.88$ compared to $7.98$, $8.71$, and $10.11$ of StyleMeUP \cite{sain2021stylemeup}, On-the-fly \cite{bhunia2020sketch}, and triplet-based baseline \cite{yu2016sketch}. Most importantly, we evaluate FG-SBIR via a human study, where our method outperforms prior arts with a massive MOS value of $4.1\pm0.5$ on the ShoeV2 dataset.}

\vspace{-0.3cm}
\begin{table}[!htbp]
\setlength{\tabcolsep}{3pt}
\renewcommand{\arraystretch}{0.8}
\centering
\caption{Results for standard FG-SBIR task.}
\vspace{-0.3cm}
\label{tab:fgsbir}
\scriptsize
\begin{tabular}{l|ccc|ccc}
\toprule
\multicolumn{1}{c|}{\multirow{3}{*}{Methods}} & \multicolumn{3}{c|}{ChairV2} & \multicolumn{3}{c}{ShoeV2} \\\cmidrule(lr){2-4}\cmidrule(lr){5-7}
& \multirow{2}{*}{Acc.@1} & \multirow{2}{*}{Acc.@5} & MOS & \multirow{2}{*}{Acc.@1} & \multirow{2}{*}{Acc.@5} & MOS \\
&  &  & $\mu\pm\sigma$ &  &  & $\mu\pm\sigma$ \\
\cmidrule(lr){1-4}\cmidrule(lr){5-7}
Triplet-SN~\cite{yu2016sketch}            & 47.4  & 71.4 & 2.6$\pm$0.4 & 28.7 & 63.5 & 2.2$\pm$0.2 \\
HOLEF-SN~\cite{song2017deep}              & 50.7  & 73.6 & 2.7$\pm$0.6 & 31.2 & 66.6 & 2.5$\pm$0.1 \\ \cmidrule(lr){1-4}\cmidrule(lr){5-7}
B-pSp                                     & 47.8  & 70.2 & 2.9$\pm$0.2 & 30.1 & 63.9 & 2.8$\pm$0.9 \\
B-pix2pix                                 & 44.2  & 66.9 & 2.8$\pm$0.7 & 26.7 & 60.2 & 2.3$\pm$0.1 \\
B-CycleGAN                                & 45.1  & 67.1 & 2.7$\pm$0.3 & 27.8 & 62.6 & 2.5$\pm$0.4 \\ \cmidrule(lr){1-4}\cmidrule(lr){5-7}
Partial-OT~\cite{chowdhury2022partially}  & 63.3  & 79.7 & 3.6$\pm$0.7 & 39.9 & 68.2 & 3.5$\pm$0.6 \\
CrossHier~\cite{sain2020cross}            & 62.4  & 79.1 & 3.5$\pm$0.3 & 36.2 & 67.8 & 3.3$\pm$0.1 \\
StyleMeUp~\cite{sain2021stylemeup}        & 62.8  & 79.6 & 3.7$\pm$0.2 & 36.4 & 68.1 & 3.3$\pm$0.3 \\
On-the-fly~\cite{bhunia2020sketch}        & 51.2  & 73.8 & 3.2$\pm$0.5 & 30.8 & 65.1 & 2.8$\pm$0.5 \\
SketchPVT~\cite{sain2023exploiting}      & 71.2  & 80.1 & 3.3$\pm$0.7 & 44.1 & 70.8 & 2.9$\pm$0.9 \\
\rowcolor{YellowGreen!40}
\textbf{\textit{Proposed}}                                  &\bf72.1  &\bf80.9 & \bf4.4$\pm$0.5 &\bf45.3  &\bf77.3 & \bf4.1$\pm$0.5 \\ \bottomrule
\end{tabular}
\end{table}
\vspace{-0.5cm}

\subsection{Sketch Abstraction Analysis}
\vspace{-0.1cm}

As there exists no  \textit{hard ground truth} or \textit{standard metric} of quantification ~\cite{muhammad2018learning,muhammad2019goal} for \emph{sketch abstraction}, we verify the generalisability of our method against different forms of abstraction (\ie, style variations~\cite{sain2021stylemeup}, method variations~\cite{alaniz2022abstracting, muhammad2018learning, muhammad2019goal}, partial/early retrieval~\cite{bhunia2020sketch, chowdhury2022partially}).

\begin{figure}[!htbp]
\vspace{-0.2cm}
    \centering
    \includegraphics[width=\columnwidth]{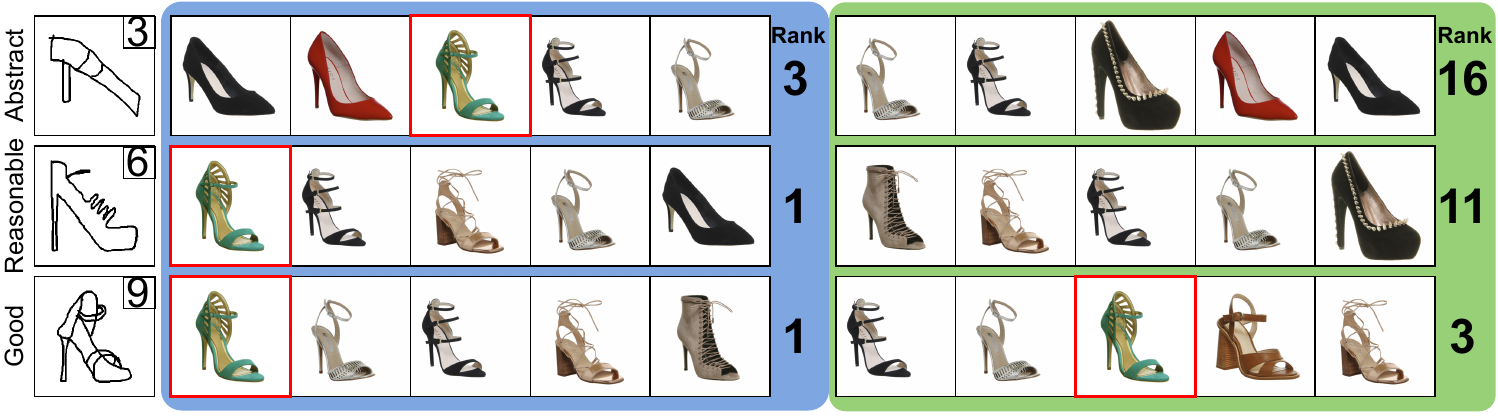}
    \vspace{-0.7cm}
    \caption{Proposed (\skyblue{blue}) method’s efficacy over Triplet-SN~\cite{yu2016sketch} (\truegreen{green}) against different sketching styles of the same shoe (\red{red} bordered). Zoom in for the best view. (More in \red{\S}~Supplementary.)}
    \label{fig:abs}
    \vspace{-0.4cm}
\end{figure}

\keypoint{Sketching Style Variations.} {Differences in abstraction-level may arise from \textit{dissimilar drawing styles} while sketching the same object~\cite{sain2021stylemeup} (See \cref{fig:abs}). The baseline Triplet-SN~\cite{yu2016sketch} fails for coarsely drawn abstract sketches due to the lack of explicit abstraction-handling. Nonetheless, our StyleGAN~\cite{karras2020analyzing}-regularised abstraction-aware model dynamically attends to varying abstraction-levels and retrieves the true match photo irrespective of the input drawing style variations. This further underpins our motivation of being truly abstraction-aware. In \cref{fig:abs}, the number beside each sketch denotes the number of row vectors from $m_s$ and $m_p$ chosen dynamically by the abstraction identification head $\mathcal{A}$. Notably, for every input, the proposed $\mathcal{A}$ module \textit{correctly} identifies the abstraction-level and selects the appropriate number of row vectors to encode the sketch.}

\keypoint{Abstraction Method Variations.} {Following \cite{alaniz2022abstracting}, we evaluate our method by abstracting the input sketches via different methods (\eg, PMN~\cite{alaniz2022abstracting}, DSA~\cite{muhammad2018learning}, GDSA~\cite{muhammad2019goal}) and budgets ($\{10,20,30,100\}\%$), corresponding to different abstraction-levels. Accordingly, the proposed method outperforms (\cref{tab:abs}) other competing SoTAs~~\cite{bhunia2020sketch,yu2016sketch} even when sketches are abstracted at $10\%$ budget, thus being in tune with our motivation of being agnostic of the abstraction methods. Please note, all competing methods in \cref{tab:abs} are trained on the original ShoeV2 sketch-image pairs.}

\vspace{-0.2cm}
\begin{table}[!htbp]
\setlength{\tabcolsep}{6pt}
\renewcommand{\arraystretch}{1}
\centering
\caption{Acc.@10 at various abstraction budgets (\%) and methods on ShoeV2. (Qualitative results in \red{\S}~Supplementary.)}
\vspace{-0.2cm}
\label{tab:abs}
\resizebox{\columnwidth}{!}{
\begin{tabular}{l|cccc|cccc|cccc}
\toprule
\multicolumn{1}{c|}{\multirow{1}{*}{Abs.}} & \multicolumn{4}{c|}{Triplet-SN~\cite{yu2016sketch}} & \multicolumn{4}{c|}{On-the-fly~\cite{bhunia2020sketch}} & \multicolumn{4}{c}{\textbf{\textit{Proposed}}} \\
               \multicolumn{1}{c|}{\multirow{1}{*}{Methods}}   & 10 & 20 & 30 & 100 & 10 & 20 & 30 & 100 & 10 & 20 & 30 & 100 \\
\cmidrule(lr){1-5}\cmidrule(lr){6-9}\cmidrule(lr){10-13}
PMN~\cite{alaniz2022abstracting}  & 9.4 & 19.3 & 33.8 & 68.4 & 10.1 & 21.6 & 40.8 & 71.9 & \cellcolor{YellowGreen!40}\textbf{51.8} & \cellcolor{YellowGreen!40}\textbf{66.1} & \cellcolor{YellowGreen!40}\textbf{68.1} & \cellcolor{YellowGreen!40}\textbf{82.3} \\
DSA~\cite{muhammad2018learning}  & 10.2 & 20.3 & 32.2 & \multirow{2}{*}{79.6} & 15.2 & 30.9 & 51.1 & \multirow{2}{*}{79.6} & \cellcolor{YellowGreen!40}\textbf{38.2} & \cellcolor{YellowGreen!40}\textbf{42.2} & \cellcolor{YellowGreen!40}\textbf{50.2} & \cellcolor{YellowGreen!40}\multirow{2}{*}{\textbf{85.1}} \\
GDSA~\cite{muhammad2019goal}  & 11.5 & 22.8 & 34.1 & & 17.1 & 31.4 & 52.9 &  & \cellcolor{YellowGreen!40}\textbf{41.7} & \cellcolor{YellowGreen!40}\textbf{48.0} & \cellcolor{YellowGreen!40}\textbf{58.3} & \cellcolor{YellowGreen!40}\multirow{-2}{*}{\textbf{85.1}}\\ \bottomrule
\end{tabular}
}
\vspace{-0.3cm}
\end{table}

\begin{figure}[!htbp]
    \centering
    \includegraphics[width=\linewidth]{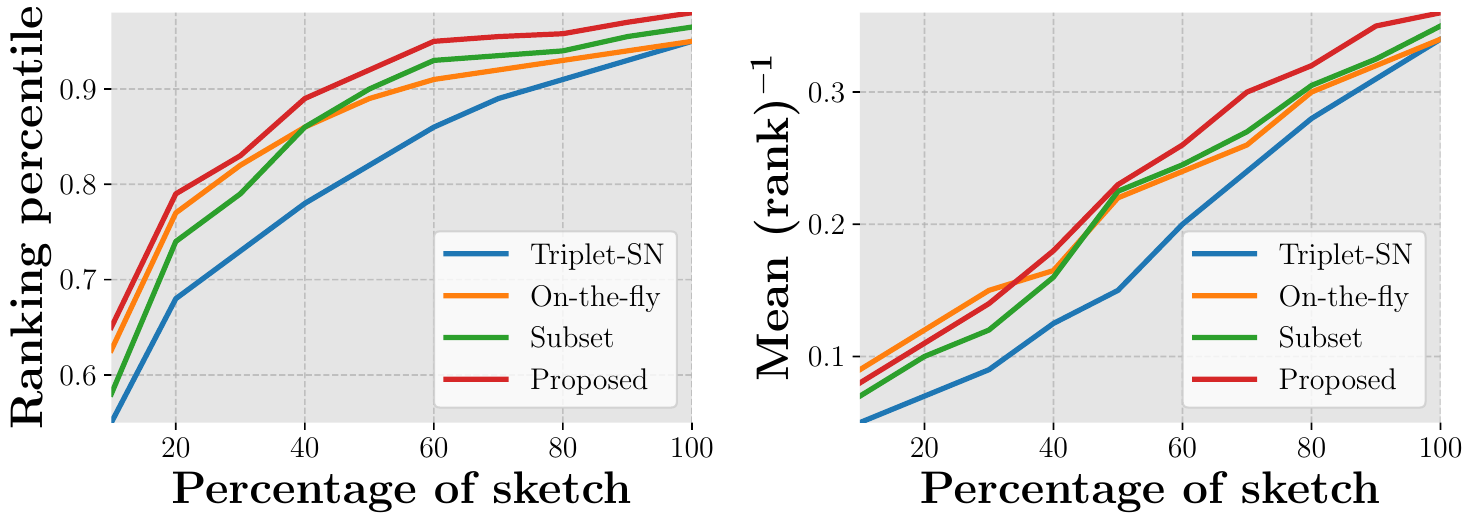}
    \vspace{-0.75cm}
    \caption{Quantitative results on ShoeV2 for early retrieval setup, visualised via the percentage of sketch. A higher area under the curve indicates better early retrieval performance.}
    \label{fig:OTF}
    \vspace{-0.8cm}
\end{figure}

\vspace{-0.1cm}
\keypoint{Partial/Early Retrieval.} {To evaluate the abstraction-aware behaviour of our method in the form of \textit{partial/early retrieval} setup, we use two curves namely \textit{(a)} ranking-percentile (m@A) and \textit{(b)} $\frac{1}{rank}$ (m@B) \vs percentage of sketch~\cite{bhunia2020sketch}. Standard methods for handling early-retrieval include Triplet-SN~\cite{yu2016sketch} with random stroke-dropping, RL fine-tuning of triplet model~\cite{bhunia2020sketch}, or RL-based salient stroke-subset selection~\cite{bhunia2022sketching}. While high-performing RL-based methods~\cite{bhunia2022sketching, bhunia2020sketch} are allegedly unstable, random stroke-dropping produces noisy gradients~\cite{bhunia2022sketching}. Contrarily, our \texttt{Acc.@q} models a similar ranking-based objective as the RL-based method \cite{bhunia2020sketch} \textit{without} extra overheads, by varying $q$ to better model partial sketches. Thanks to \texttt{Acc.@q}, our model outperforms (\cref{fig:OTF}) On-the-fly~\cite{bhunia2020sketch}, and Subset~\cite{bhunia2022sketching}. Overall, our method achieves a higher m@A (m@B) of $86.22$ ($22.30$) as compared to $85.38$ ($21.24$) and $85.78$ ($21.1$) claimed in~\cite{bhunia2020sketch} and~\cite{bhunia2022sketching} respectively.}

\vspace{-0.2cm}

\subsection{Ablation}
\vspace{-0.1cm}
\noindent\textbf{[i] Contribution of \texttt{Acc.@q} and Reconstruction Loss.} To estimate the efficacy of the proposed surrogate loss function, we replace \texttt{Acc.@q} with a standard triplet-based \cite{yu2016sketch} objective. Replacing \texttt{Acc.@q} causes a sharp MOS value decrease of $1.7\pm0.1$ on ShoeV2 (\cref{tab:ablation}). We posit that this drop is due to the absence of the test-time metric like ranking guidance provided by the surrogate loss. Additionally, replacing \texttt{Acc.@q} with SoTA surrogate losses like SoDeep \cite{engilberge2019sodeep} and SmoothAP \cite{brown2020smooth} plummets MOS value by $1.1\pm0.2$ and $0.9\pm0.1$ respectively on ShoeV2. Moreover, as evident from the \textbf{w/o recons.\ loss} result, exclusion of the reconstruction loss destabilises the system, causing an additional Acc.@1 drop of $10.1\%$ on the ShoeV2 dataset. Furthermore, fine-tuning the StyleGAN model (instead of frozen), results in lower Acc.@1 ($30.2\%$ on ShoeV2), likely because fine-tuning sacrifices the latent space disentanglement.

\vspace{-0.3cm}
\begin{table}[!hbt]
\setlength{\tabcolsep}{2pt}
\renewcommand{\arraystretch}{1}
\centering
\caption{{Ablation on design.}}
\vspace{-0.35cm}
\label{tab:ablation}
\scriptsize
\resizebox{\columnwidth}{!}{
\begin{tabular}{l|ccc|ccc}
\toprule
\multicolumn{1}{c|}{\multirow{3}{*}{Methods}} & \multicolumn{3}{c|}{ChairV2} & \multicolumn{3}{c}{ShoeV2} \\\cmidrule(lr){2-4}\cmidrule(lr){5-7}
& \multirow{2}{*}{Acc.@1} & \multirow{2}{*}{Acc.@5} & MOS & \multirow{2}{*}{Acc.@1} & \multirow{2}{*}{Acc.@5} & MOS \\
&  &  & $\mu\pm\sigma$ &  &  & $\mu\pm\sigma$ \\
\cmidrule(lr){1-4}\cmidrule(lr){5-7}
w/o abs.\ identification     & 59.2  & 71.9 & 3.0$\pm$0.1 & 38.1 & 69.2 & 2.5$\pm$0.3 \\
w/o $\texttt{Acc.@q}$ loss   & 58.5  & 70.1 & 2.8$\pm$0.3 & 37.1 & 68.1 & 2.4$\pm$0.4 \\
w/o recons.\ loss            & 56.7  & 69.6 & 2.3$\pm$0.2 & 35.2 & 66.1 & 2.0$\pm$0.1 \\ \cmidrule(lr){1-4}\cmidrule(lr){5-7}
Ours+SoDeep~\cite{engilberge2019sodeep}        & 60.3  & 74.1 & 2.7$\pm$0.4 & 39.6 & 71.3 & 3.0$\pm$0.3 \\
Ours+SmoothAP~\cite{brown2020smooth}           & 61.7  & 74.9 & 3.1$\pm$0.2 & 40.3 & 73.9 & 3.2$\pm$0.4 \\
\rowcolor{YellowGreen!40}
\textbf{\textit{Ours-full}}                    & \bf72.1 & \bf80.9 & \bf4.2$\pm$ 0.3 & \bf45.3 & \bf77.3 & \bf4.1$\pm$0.5 \\

\textit{\graytext{Avg. Improvement}}                    & \textit{\graytext{+11.1}} & \textit{\graytext{+6.4}} & \textit{\graytext{+1.3$\pm$ 0.0}} & \textit{\graytext{+5.3}} & \textit{\graytext{+4.7}} & \textit{\graytext{+1.0$\pm$0.1}} \\ \bottomrule
\end{tabular}
}
\vspace{-0.3cm}
\end{table}

\noindent\textbf{[ii] Importance of Abstraction Identification (Eq.\ \ref{eq:abs_identification}).} To illustrate the relevance of our abstraction identification head, instead of three-stage rendering, we rasterise the full sketch vector at once, prior to training, thus using the entire feature matrix embedding of size $\mathbb{R}^{9\times d}$ regardless of the input abstraction. Although less reflected in Acc.@1, a significant MOS drop of $1.6\pm0.2$ on ShoeV2 (\cref{tab:ablation}) in case of \textbf{w/o abs.\ identification} signifies that sketch abstraction-modelling is incomplete without learning to predict the input abstraction-level via partial rendering.

\noindent\textbf{[iii] Efficiency Analysis.} As the gallery photo features are \textit{pre-computed} during inference, the only difference in efficiency between a standard VGG-16 triplet model~\cite{yu2016sketch} and ours will arise from the additional \textit{sketch feature matrix embedding network} $\mathcal{E}_s$. While the former takes $40.18$G FLOPs to process a query-sketch of size $256\times 256$, we use $40.20$G FLOPs, which is \textit{only} $0.05\%$ higher yet boosts accuracy by $16.6\%$. Secondly, unlike the $\mathbb{R}^{d}$ feature \textit{vector} of triplet-based models~\cite{yu2016sketch}, we use a $\mathbb{R}^{9\times d}$ embedding \textit{matrix} for distance calculation during inference. Thus, \textit{theoretically}, a triplet model should be $9\times$ faster. However, in a practical FG-SBIR inference pipeline of \textit{feature extraction} followed by \textit{distance calculation} and \textit{ranking}, our matrix distance calculation takes only a \textit{negligible} $0.04\%$ ($0.03$ms) out of a total of $77.1$ms, on a $12$ GB Nvidia $2080$ Ti GPU. Besides, as the pre-trained StyleGAN is \textit{used only during training}, its architectural complexities \textit{do not} affect the inference.

\noindent\textbf{[iv] Latent Sub-grouping.} We segregate the structural latent vectors into three sub-groups, each with a set of \textit{three} latent vectors $\{w^+_{1\shortrightarrow3}, w^+_{4\shortrightarrow6}, w^+_{7\shortrightarrow9}\}$ to model varying the sketch abstraction-levels (\cref{sec:method}). Testing with most such group combinations, we find that the $\{a_{c},a_{m},a_{f}\}=\{3,3,3\}$ combination is the most \textit{generalised} and gives the \textit{optimum} Acc.@1 ($45.3\%$) compared to others (\eg, $\{2, 3, 4\}$--$39.8\%$, $\{2, 2, 5\}$--$33.6\%$, $\{1, 2, 6\}$--$28.9\%$, etc.).

\vspace{-0.1cm}
\subsection{Extension to Forensic Sketch-Photo Matching}
\vspace{-0.1cm}
{Forensic sketch-photo matching is among the many practical applications of SBIR, where the system uses a forensic facial sketch as a query to recognise a person within a vast gallery (closed-set) of images~\cite{cho2020relational, bhatt2012memetically}. {Forensic-sketching} depicts varying levels of abstraction based on the \textit{drawing skill} (or the lack of it) of the artist~\cite{deng2019residual, bhatt2012memetically}. Here we explore the potential of the information-rich latent space of StyleGAN~\cite{karras2020analyzing} for solving the challenging problem of forensic sketch-photo recognition. Using the large-scale FFHQ~\cite{karras2019style} pre-trained StyleGAN model, we learn the backbone and the feature matrix encoder leveraging the IIIT-D viewed-sketch dataset (\cref{fig:iitd})~\cite{bhatt2012memetically}, which includes \textit{only one} sketch-photo pair for each of $238$ different human subjects. Besides the standard Acc.@1, following the evaluation paradigm of~\cite{wu2018coupled}, we also use the verification rate (VR) @ false acceptance rate (FAR) = $1\%$ metric. Compared to other SoTA methods with complex architectures~\cite{deng2019residual,wu2018coupled,cho2020relational}, handling data scarcity by training their models on the \textit{much larger} CUFSF~\cite{zhang2011coupled} dataset, our method despite being supervised with \textit{only one} sketch-photo pair per subject, surpasses (\cref{tab:faceRec}) them with an average Acc.@1 gain of $10.77\%$.}

\begin{table}[!htbp]
\vspace{-0.2cm}
  \setlength{\tabcolsep}{2pt}
  \renewcommand{\arraystretch}{0.8}
  \scriptsize
    \begin{minipage}{0.48\linewidth}
    \centering
    \includegraphics[width=35mm]{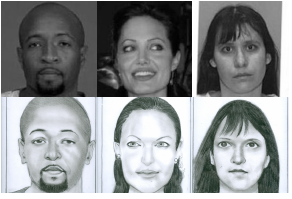}
    \vspace{-0.30cm}
    \captionof{figure}{Sketch-photo samples from the IIIT-D~\cite{bhatt2012memetically} dataset.}
    \label{fig:iitd}
  \end{minipage}\hfill
  \begin{minipage}{0.5\linewidth}
    \caption{Results for forensic sketch-photo recognition.}
    \label{tab:faceRec}
    \centering
    \vspace{-0.35cm}
    \begin{tabular}{l|cc}
      \toprule
      \multicolumn{1}{c|}{\multirow{2}{*}{Methods}} & \multirow{2}{*}{Acc.@1} & VR \\
                               &                         & @FAR=1\% \\
      \midrule
      RCN~\cite{deng2019residual}              & 63.83    & 90.12  \\
      Deep-Face~\cite{parkhi2015deep}          & 80.89    & 72.08  \\
      Center-Loss~\cite{wen2016discriminative} & 84.07    & 76.20  \\
      CDL~\cite{wu2018coupled}                 & 85.35    & 82.52  \\
      RGM~\cite{cho2020relational}             & 88.94    & \bf97.87  \\
      \rowcolor{YellowGreen!40}
      \textbf{\textit{Proposed}}                         & \bf91.39 & 97.38  \\\bottomrule
    \end{tabular}
  \end{minipage}
\end{table}

 \vspace{-0.63cm}
\section{Conclusion and Future Works}
\vspace{-0.1cm}
The proposed generation-guided FG-SBIR method along with a differentiable surrogate ranking loss explicitly handles freehand sketch abstraction, allowing it to retrieve accurate images even from highly \textit{abstract} sketches. Extensive evaluations outperforming existing SoTAs depict the efficacy of StyleGAN's disentangled latent space in abstraction-aware fine-grained retrieval tasks. Our method could further be extended to scene-level SBIR with scene-level GANs \cite{sauer2022stylegan, kang2023scaling}. Moreover, the synergy between freehand sketches and pre-trained GAN could pave the way for different weakly-supervised sketch-based applications like object co-part segmentation \cite{hung2019scops}, image editing \cite{richardson2021encoding}, etc.

{
    \small
    \bibliographystyle{ieeenat_fullname}
    \bibliography{arxiv}
}
\clearpage

\onecolumn{
\centering
\title{\Large{\textbf{Supplementary material for\\ How to Handle \emph{Sketch-Abstraction} in Sketch-Based Image Retrieval?}}}
\vspace{0.4cm}

\author{\MYhref[cvprblue]{https://subhadeepkoley.github.io}{Subhadeep Koley}\textsuperscript{1,2} \hspace{.2cm} \MYhref[cvprblue]{https://ayankumarbhunia.github.io}{Ayan Kumar Bhunia}\textsuperscript{1} \hspace{.2cm} \MYhref[cvprblue]{https://aneeshan95.github.io}{Aneeshan Sain}\textsuperscript{1} \hspace{.2cm}  \MYhref[cvprblue]{https://www.pinakinathc.me}{Pinaki Nath Chowdhury}\textsuperscript{1} \\ \MYhref[cvprblue]{https://www.surrey.ac.uk/people/tao-xiang}{Tao Xiang}\textsuperscript{1,2} \hspace{.2cm} \MYhref[cvprblue]{https://www.surrey.ac.uk/people/yi-zhe-song}{Yi-Zhe Song}\textsuperscript{1,2} \\
\textsuperscript{1}SketchX, CVSSP, University of Surrey, United Kingdom.  \\
\textsuperscript{2}iFlyTek-Surrey Joint Research Centre on Artificial Intelligence.\\
{\tt\small \{s.koley, a.bhunia, a.sain, p.chowdhury, t.xiang, y.song\}@surrey.ac.uk}
}

}

\maketitle

\section*{A. Additional Qualitative Results}
\begin{figure}[!htbp]
    \centering
    \includegraphics[width=0.9\linewidth]{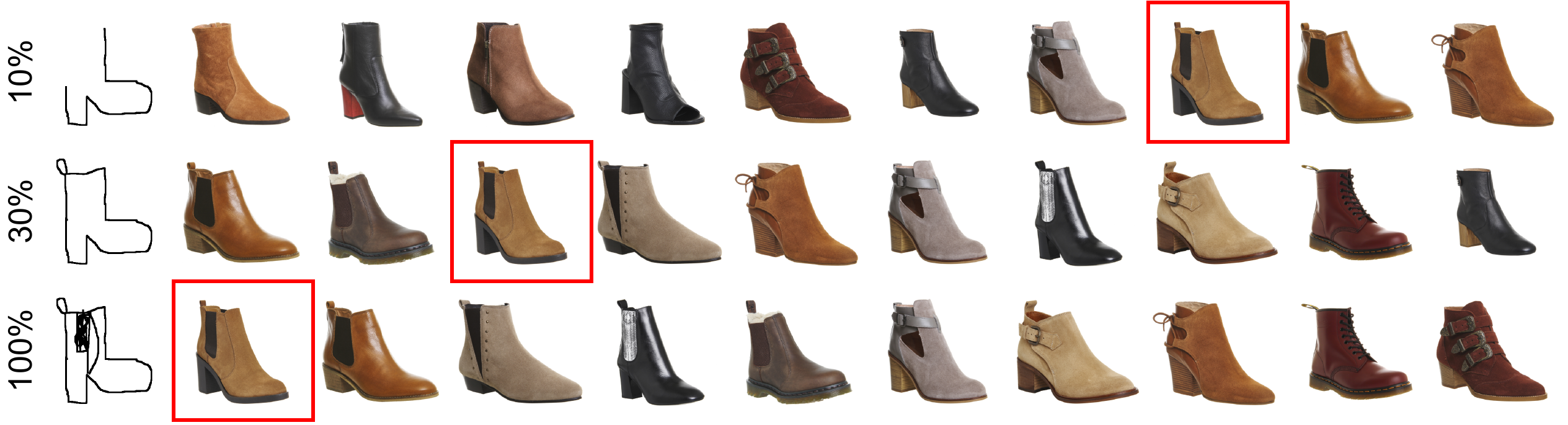}
    \vspace{-0.3cm}
    \caption{Top-10 retrieved images for inputs abstracted (by~\cite{muhammad2019goal}) at different budgets ($10\%, 30\%, 100\%$). Paired photo is \red{red} bordered.}
    \label{fig:budget_1}
\end{figure}

\begin{figure}[!htbp]
    \centering
    \includegraphics[width=0.9\linewidth]{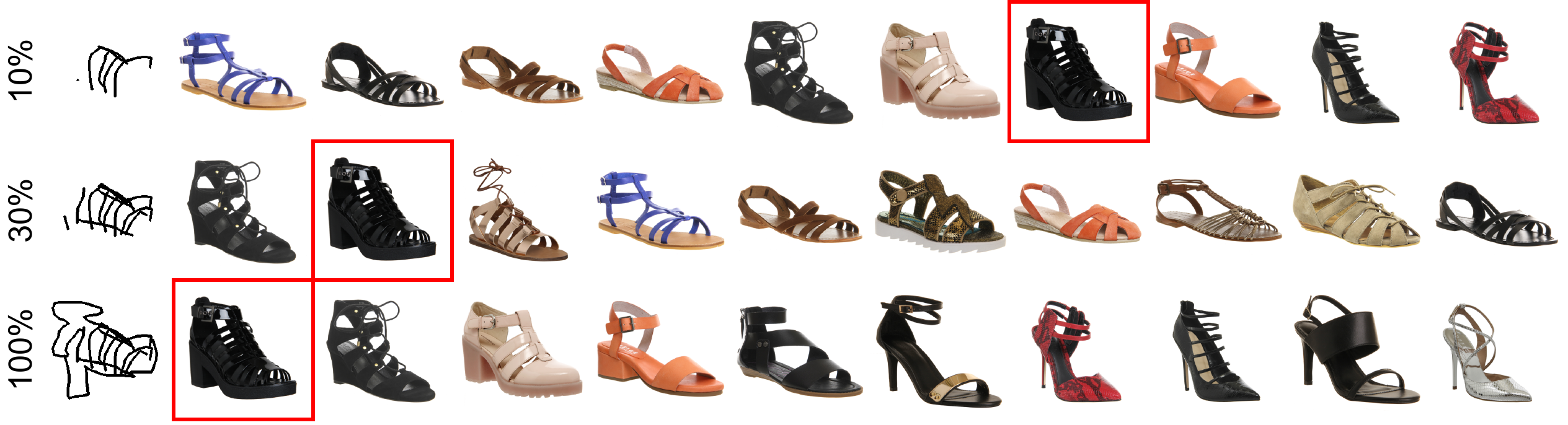}
    \vspace{-0.3cm}
    \caption{Top-10 retrieved images for inputs abstracted (by~\cite{muhammad2019goal}) at different budgets ($10\%, 30\%, 100\%$). Paired photo is \red{red} bordered.}
    \label{fig:budget_2}
\end{figure}

\begin{figure}[!htbp]
    \centering
    \includegraphics[width=0.85\linewidth]{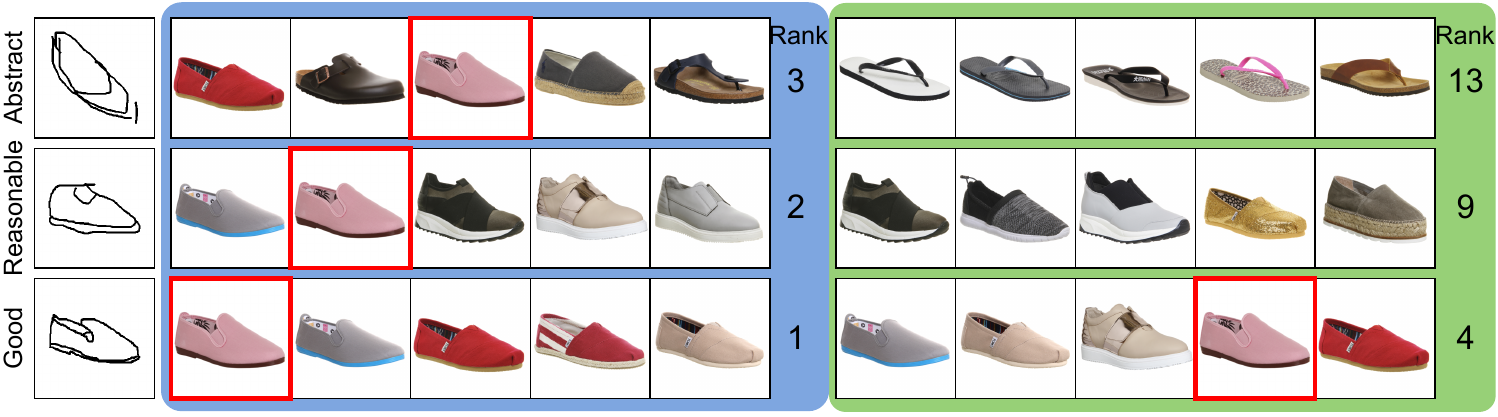}
    \vspace{-0.3cm}
    \caption{Proposed (\skyblue{blue}) method’s efficacy over Triplet-SN~\cite{yu2016sketch} (\truegreen{green}) against different sketching styles of the same shoe (\red{red} bordered).}
    \label{fig:GRA_1}
\end{figure}

\begin{figure}[!htbp]
    \centering
    \includegraphics[width=0.85\linewidth]{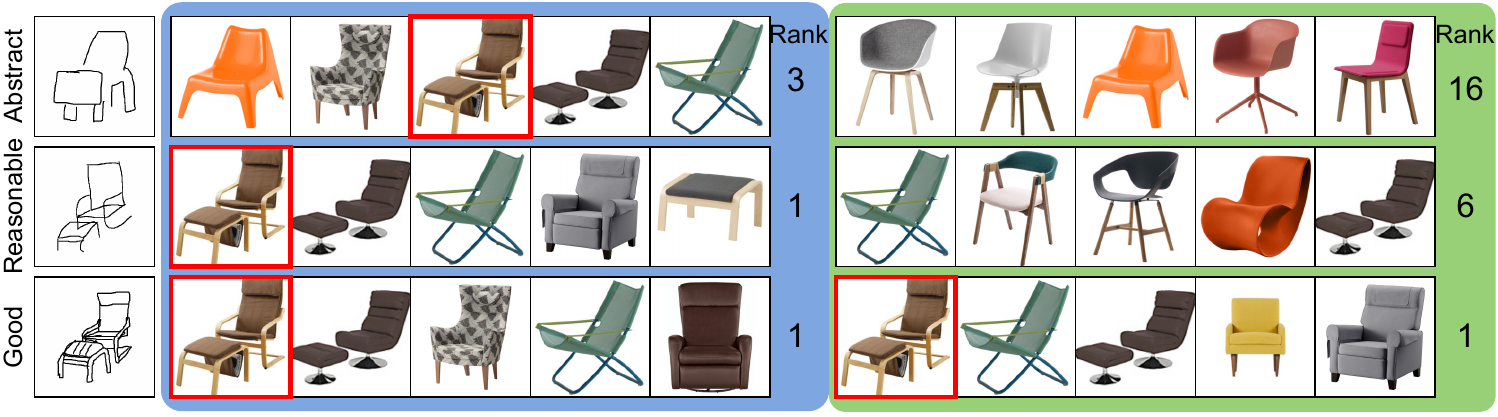}
    \vspace{-0.3cm}
    \caption{Proposed (\skyblue{blue}) method’s efficacy over Triplet-SN~\cite{yu2016sketch} (\truegreen{green}) against different sketching styles of the same chair (\red{red} bordered).}
    \label{fig:GRA_2}
\end{figure}

\begin{figure}[!htbp]
    \centering
    \includegraphics[width=0.80\linewidth]{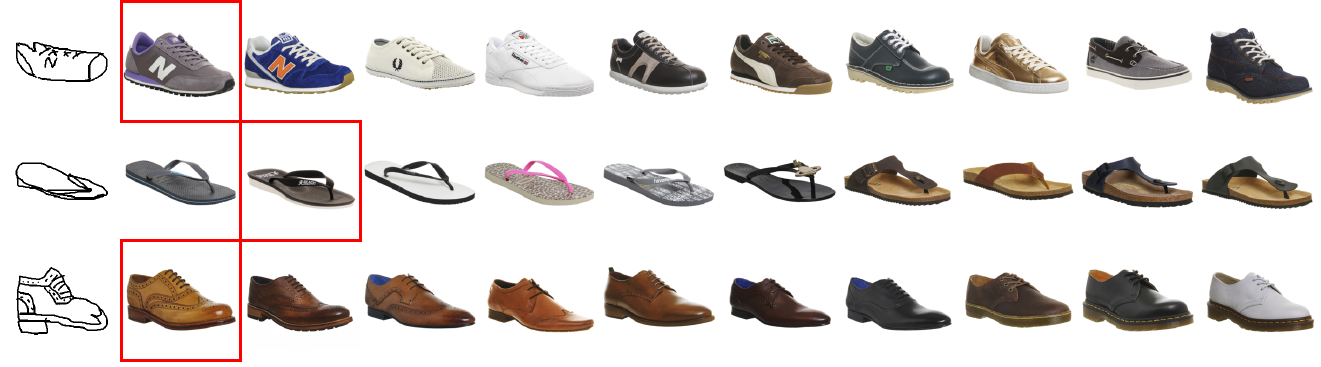}
    \vspace{-0.3cm}
    \caption{Top-10 qualitative retrieval results of the proposed method on sketches from ShoeV2 dataset~\cite{bhunia2020sketch, song2018learning}. Paired photo is \red{red} bordered.}
    \label{fig:qual_1}
\end{figure}

\begin{figure}[!htbp]
    \centering
    \includegraphics[width=0.80\linewidth]{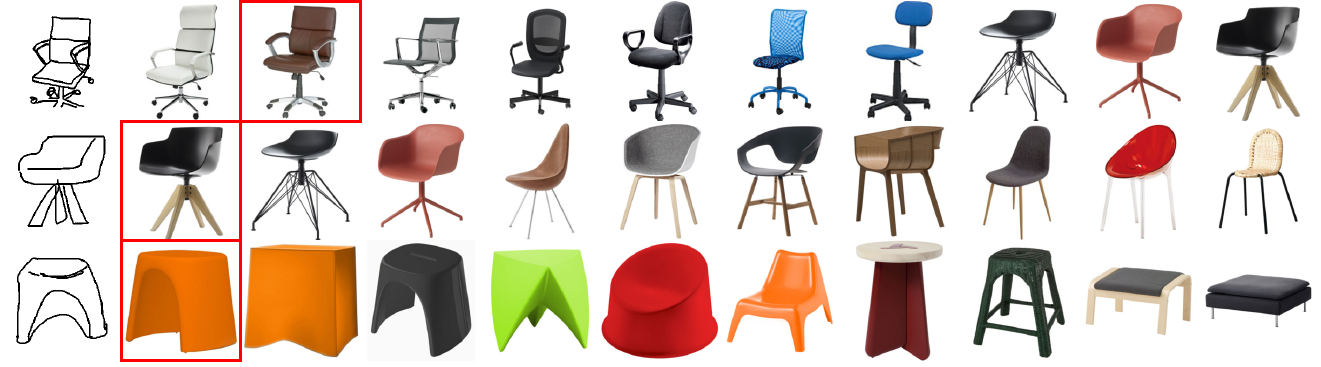}
    \vspace{-0.3cm}
    \caption{Top-10 qualitative retrieval results of the proposed method on sketches from ChairV2 dataset~\cite{bhunia2020sketch, song2018learning}. Paired photo is \red{red} bordered.}
    \label{fig:qual_2}
\end{figure}

\vspace{-0.2cm}
Figs.\ \ref{fig:budget_1}-\ref{fig:qual_2} depict additional qualitative retrieval results for various retrieval scenarios and datasets~\cite{bhunia2020sketch, song2018learning} using our framework. To delineate the abstraction-agnostic behaviour of our method, we abstracted the input sketches using the GDSA~\cite{muhammad2019goal} method at different abstraction budgets ($\{10, 30, 100\}\%$). \cref{fig:budget_1} and \cref{fig:budget_2} show how our method reasonably retrieves the ground truth paired photo even in the case of \textit{extreme} abstraction of $10\%$.

\vspace{-0.3cm}

\cref{fig:GRA_1} and \cref{fig:GRA_2} qualitatively depict our method’s efficacy over Triplet-SN~\cite{yu2016sketch} for the case of different sketching styles (\ie, good, reasonable, and abstract) of the same object. It is evident from \cref{fig:GRA_1} and \cref{fig:GRA_2} that the proposed method equipped with dynamic abstraction identification surpasses SoTA Triplet-SN~\cite{yu2016sketch} in every case.
\clearpage

\vspace{-0.8cm}

\vspace{-0.8cm}
\vspace{-0.3cm}

Finally, \cref{fig:qual_1} and \cref{fig:qual_2} show qualitative retrieval results of the proposed method on sketches from ShoeV2~\cite{bhunia2020sketch, song2018learning} and ChairV2~\cite{bhunia2020sketch, song2018learning} datasets. It is worth noticing in \cref{fig:qual_1} and \cref{fig:qual_2} how the retrieved images \textit{transition smoothly} from rank-1 to rank-10. This importantly ensures that most of the images retrieved by the proposed method are \textit{semantically relevant} and \textit{correspond to the input sketch}. We posit that this behaviour is driven by the regularisation provided by the disentangled and smooth latent space of StyleGAN~\cite{karras2020analyzing}.

\vspace{-0.3cm}
\vspace{-0.8cm}

\clearpage
\section*{B. Quantitative Analysis of Dynamic Structural Latent Code Selection}
\vspace{-0.2cm}
To quantitatively demonstrate the relevance of the proposed dynamic structural latent code selection through the abstraction identification head ($\mathcal{A}$), we perform a few experiments. Here, instead of the automatic prediction of embedding matrix dimension via the $\mathcal{A}$ module, we force the system to always use either $3$, $6$, or $9$ structural latent codes \textit{regardless} of the input abstraction, thus resulting in a feature embedding matrix of size $\mathbb{R}^{3\times d}$, $\mathbb{R}^{6\times d}$, or $\mathbb{R}^{9\times d}$ respectively. Additionally, in another paradigm we enforce the model to \textit{randomly} select between the feature embedding matrix of size $\mathbb{R}^{3\times d}$, $\mathbb{R}^{6\times d}$, or $\mathbb{R}^{9\times d}$. Experimental results in \cref{tab:latent_sel} depict how the accuracy falls drastically in cases of fixed or random latent selection. On the other hand, the proposed method equipped with \textit{dynamic abstraction-modelling} outperforms them all with an Acc.@1 of $45.3\% (72.1\%)$ on ShoeV2 (ChairV2) dataset.

\vspace{-0.3cm}
\begin{table}[!hbt]
\centering
\caption{Quantitative analysis of dynamic latent selection.}
\vspace{-0.3cm}
\label{tab:latent_sel}
\scriptsize
\begin{tabular}{l|cc|cc}
\toprule
\multicolumn{1}{c|}{\multirow{2}{*}{Embedding matrix dimension}} & \multicolumn{2}{c|}{ChairV2} & \multicolumn{2}{c}{ShoeV2} \\\cmidrule(lr){2-3} \cmidrule(lr){4-5}
& Acc.@1 & Acc.@5 & Acc.@1 & Acc.@5 \\
\cmidrule(lr){1-3}\cmidrule(lr){4-5}
$\mathbb{R}^{3\times d}$     & 23.6  & 42.8 & 18.9 & 31.7 \\
$\mathbb{R}^{6\times d}$     & 44.7  & 53.6 & 29.2 & 49.8 \\
$\mathbb{R}^{9\times d}$     & 58.5  & 70.1 & 37.1 & 68.1 \\
Random $(\mathbb{R}^{3\times d}$ / $\mathbb{R}^{6\times d}$ / $\mathbb{R}^{9\times d})$       & 45.4  & 55.1 & 30.5 & 51.3 \\
\rowcolor{YellowGreen!40}
\textbf{\textit{Ours-full}}         & \textbf{72.1}  & \textbf{80.9} & \textbf{45.3} & \textbf{77.3} \\
\textit{\graytext{Avg.\ Improvement}}               & \textit{\graytext{+29.0}}  & \textit{\graytext{+25.5}} & \textit{\graytext{+16.3}} & \textit{\graytext{+27.0}} \\
\bottomrule
\end{tabular}
\vspace{-0.65cm}
\end{table}

\section*{C. Utility of Abstraction-aware Feature Matrix Embedding}
\begin{figure}[!htbp]
\centering
    \centering
    \includegraphics[width=0.3\columnwidth]{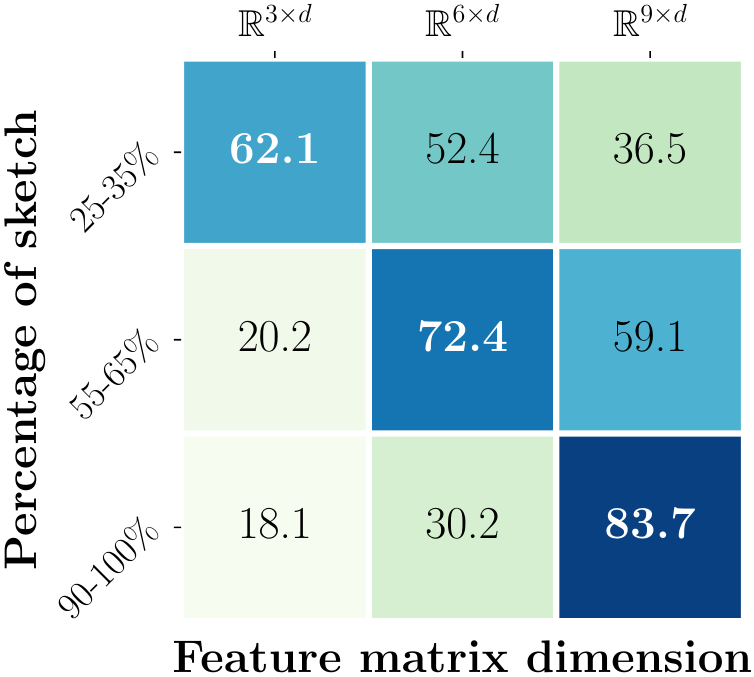}
    \vspace{-0.3cm}
    \caption{Acc.@10 comparison at different feature matrix dimension and sketch completion levels.}
    \vspace{-0.2cm}
    \label{fig:confusion}
\end{figure}

\vspace{-0.3cm}
Recent literature~\cite{karras2019style, karras2020analyzing, patashnik2021styleclip, richardson2021encoding, yang2021semantic} motivates us to exploit the abstraction hierarchy present in the StyleGAN~\cite{karras2019style} latent matrix. To justify the same, we experiment by rendering the ShoeV2 test set sketches at different stages ($25$-$35\%$, $55$-$65\%$, \& $90$-$100\%$) to represent three abstraction levels and forcing the model to calculate the distance with $\mathbb{R}^{3\times d}$, $\mathbb{R}^{6\times d}$, and $\mathbb{R}^{9\times d}$ dimensional feature matrices \textit{per level}. The resultant plot (\cref{fig:confusion}) shows how the proposed matrix embedding achieves \textit{optimum} Acc.@10 for each abstraction level when the distance is calculated with the \textit{corresponding} matrix dimension by \textit{traversing} the rows of the matrix embedding. This underpins our hypothesis that the feature matrix embedding can efficiently \textit{accommodate} different abstraction levels.

\vspace{-0.3cm}

\vspace{-0.5cm}
\section*{D. Choice of Backbone and Additional StyleGAN-based Baselines}
\vspace{-0.2cm}
In our approach, the pre-trained StyleGAN~\cite{karras2020analyzing} is used \textit{only} during training. During inference, we \textit{discard} it and use the trained VGG-16~\cite{simonyan2014very} feature extractor ($\mathcal{F}$), along with two sketch and photo-specific feature-matrix embedding networks $\mathcal{E}_s$ and $\mathcal{E}_p$ to calculate the matrix embeddings. However, following SoTAs~\cite{bhunia2020sketch,sain2021stylemeup,bhunia2022sketching,chowdhury2022partially,sain2020cross}, training an ImageNet-pretrained Inception-V3~\cite{szegedy2016rethinking} feature extractor ($\mathcal{F}$) we get a competitive Acc.@1 of $47.1 (71.4)\%$ on ShoeV2 (ChairV2), thus validating our comparisons. On the other hand, to justify the proposed usage of a pre-trained StyleGAN~\cite{karras2020analyzing} and for a fairer comparison, we amend existing SoTAs \cite{yu2016sketch,song2017deep,chowdhury2022partially,sain2020cross,sain2021stylemeup} with an additional StyleGAN-based regularisation. Here, given the respective SoTA backbones $\mathcal{F}(I)\in \mathbb{R}^{h\times w\times d}$, we employ $14$ individual stride-two convolution blocks with LeakyReLU~\cite{xu2015empirical} applied over $\mathcal{F}(I)$ to convert $\mathbb{R}^{h\times w\times d}\rightarrow\mathbb{R}^{14\times d}$. This $\mathbb{R}^{14\times d}$ code upon passing through a pre-trained StyleGAN generates an image $\hat{p}$ for both photo ($\hat{p}_p$) and sketch ($\hat{p}_s$) branches, which we utilise to impose an additional reconstruction objective ($\mathcal{L}_{\text{recons}}$) apart from their respective losses.

\vspace{-0.5cm}
\begin{equation}
\label{eq:l_rec}
\mathcal{L}_{\text{recons}} =   \left \| p -  \hat{p}_s\right \|_2 +  \left \| p - \hat{p}_p \right \|_2
\end{equation}
\vspace{-0.5cm}

\noindent This importantly ensures that the pre-trained StyleGAN's knowledge is distilled into the SoTA frameworks~\cite{yu2016sketch,song2017deep,chowdhury2022partially,sain2020cross,sain2021stylemeup}. Although this trick boosts SoTA performance up to a certain extent (\cref{tab:sg_baselines}), the proposed method surpasses them with an average Acc.@1 of $5.6\% (12.2\%)$ on ShoeV2 (ChairV2). This further explains how the naive adaptation of StyleGAN fails to efficiently utilise the rich information residing in a pre-trained StyleGAN's latent space.

\vspace{-0.2cm}
\begin{table}[!hbt]
\centering
\caption{Quantitative analysis of StyleGAN-based regularisation.}
\vspace{-0.2cm}
\label{tab:sg_baselines}
\scriptsize
\begin{tabular}{l|cc|cc}
\toprule
\multicolumn{1}{c|}{\multirow{2}{*}{Methods}} & \multicolumn{2}{c|}{ChairV2} & \multicolumn{2}{c}{ShoeV2} \\\cmidrule(lr){2-3} \cmidrule(lr){4-5}
& Acc.@1 & Acc.@5 & Acc.@1 & Acc.@5 \\
\cmidrule(lr){1-3}\cmidrule(lr){4-5}
Triplet-SN~\cite{yu2016sketch} + $\mathcal{L}_{\text{recons}}$              & 50.2  & 75.4 & 33.3 & 68.2 \\
HOLEF-SN~\cite{song2017deep} + $\mathcal{L}_{\text{recons}}$                & 52.6  & 77.1 & 34.8 & 69.9 \\
Partial-OT~\cite{chowdhury2022partially} + $\mathcal{L}_{\text{recons}}$    & 66.2  & 82.8 & 43.1 & 71.8 \\
CrossHier~\cite{sain2020cross} + $\mathcal{L}_{\text{recons}}$              & 64.9  & 80.9 & 42.8 & 72.3 \\
StyleMeUp~\cite{sain2021stylemeup} + $\mathcal{L}_{\text{recons}}$          & 65.4  & 80.1 & 44.6 & 73.5 \\

\rowcolor{YellowGreen!40}
\textbf{\textit{Ours-full}}                    & \textbf{72.1}  & \textbf{80.9} & \textbf{45.3} & \textbf{77.3} \\
\textit{\graytext{Avg.\ Improvement}}               & \textit{\graytext{+12.2}}  & \textit{\graytext{+1.6}} & \textit{\graytext{+5.6}} & \textit{\graytext{+6.1}} \\
\bottomrule
\end{tabular}
\end{table}

\vspace{-0.5cm}
\section*{E. Comparison with Other Surrogate Losses}
\vspace{-0.2cm}
Cross-modal retrieval is typically evaluated on three metrics -- accuracy, precision, and recall~\cite{sain2021stylemeup,bhunia2022sketching}. While \textit{precision} and \textit{recall} measure \textit{how well} or \textit{how many times} the model detected a certain \textit{category} respectively, \textit{accuracy} indicates the overall model performance irrespective of the category, thus making it the standard metric for instance-level fine-grained retrieval tasks~\cite{wei2021fine}. Existing surrogate losses~\cite{brown2020smooth,patel2022recall} mostly optimise category-level metrics (\eg, precision~\cite{brown2020smooth} or recall~\cite{patel2022recall}), rendering them sub-optimal for our \textit{fine-grained} setting. On the other hand, Engilberge et al.~\cite{engilberge2019sodeep} proposed an LSTM-based network to learn ranking loss surrogates, but its adaptation has been limited in the consequent literature due to the alleged slow training~\cite{patel2022recall}. Although SmoothAP~\cite{brown2020smooth} and Recall@k~\cite{patel2022recall} have shown promising results in fine-grained datasets like INaturalist~\cite{van2018inaturalist} and VehicleID~\cite{van2018inaturalist}, their off-the-shelf adaptation in our cross-modal fine-grained scenario produces sub-optimal Acc.@1 of $40.3\%$ and $39.5\%$ respectively in ShoeV2. On the other hand, the proposed \texttt{Acc.@q} loss being tailored for smooth approximation of the instance-level retrieval metric (\ie, accuracy), outperforms existing SoTAs by a significant margin. More importantly, the parametric design of our \texttt{Acc.@q} loss allows us to use \textit{different} variants (by changing $q=1/5/10$) of the same loss to tackle \textit{different} abstraction levels, which in turn provide better \textit{retrieval granularity}.

\section*{F. Details on Human Study}
\begin{figure}[!htbp]
    \centering
    \includegraphics[width=0.73\linewidth]{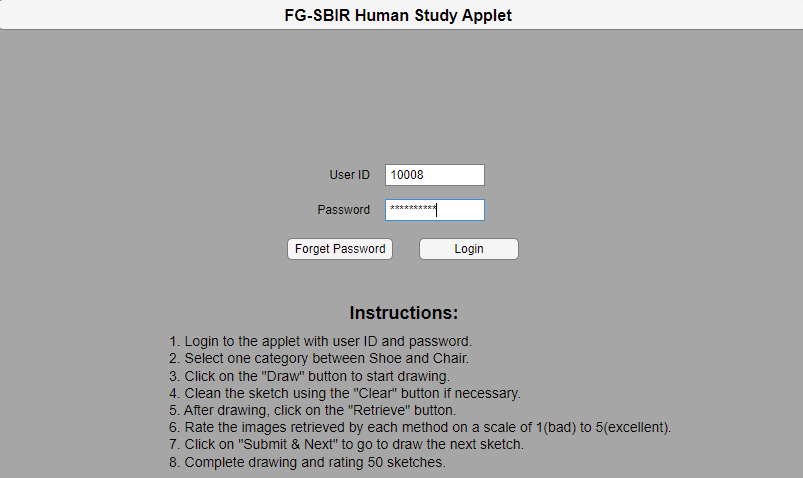}
    \vspace{-0.3cm}
    \caption{Login UI of the FG-SBIR human study applet}
    \label{fig:mos_login}
\end{figure}

\begin{figure}[!htbp]
    \centering
    \includegraphics[width=0.73\linewidth]{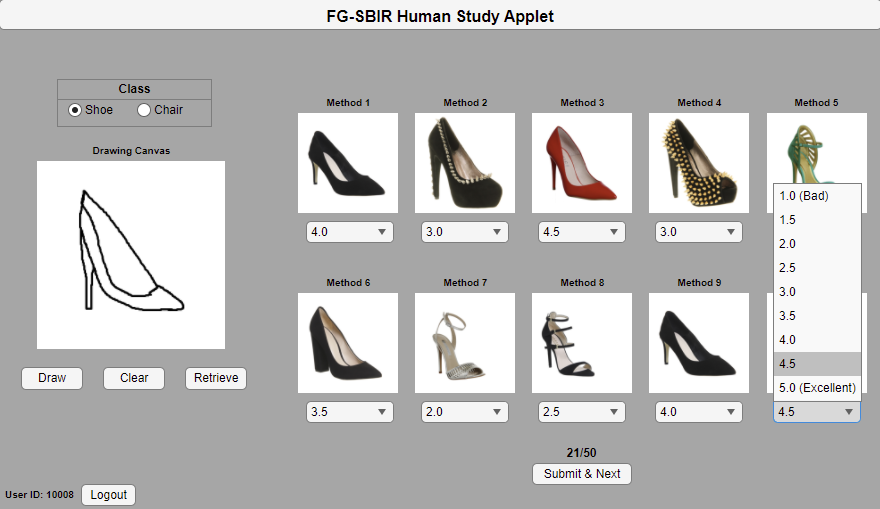}
    \vspace{-0.3cm}
    \caption{Scoring UI of the FG-SBIR human study applet}
    \label{fig:mos_scoring}
\end{figure}

\vspace{-0.2cm}
\cref{fig:mos_login} and \cref{fig:mos_scoring} depict various UIs of the applet used to collect Mean Opinion Scores (MOS)~\cite{huynh2010study} through a human study. After logging into the system, the participant first selects the category (\ie, shoe or chair) of which class he/she wants to draw a sketch. Next, the user clicks on the ``Draw'' button to activate the drawing tool and starts drawing. Upon finishing, the participant clicks on the ``Retrieve'' button to view the images retrieved by all competing methods. The user rates every retrieved photo and clicks on ``Submit \& Next'' to continue. We further sub-divide the MOS value levels ($1$(bad)$\rightarrow$$5$(excellent)) into nine discreet levels (\eg, $\{1,1.5,2,2.5,3,3.5,4,4.5,5\}$)~\cite{huynh2010study} for brevity and ease of rating. We \textit{purposefully} anonymise the method names to prevent the rating from being influenced by the participant's prior knowledge of the literature.

\vspace{-0.2cm}

\end{document}